\documentclass[twoside,11pt]{article}
\usepackage[preprint]{jmlr2e}

\usepackage{amsmath}
\usepackage{booktabs}
\usepackage{array}
\usepackage{float}
\usepackage{microtype}
\usepackage[section]{placeins}
\hypersetup{hidelinks,hypertexnames=false}
\ShortHeadings{Orthogonality Beyond Sphere Optimization}{Hu, Zhang, Yu, Fredrikson, and Hu}
\firstpageno{1}

\begin{document}

\title{Do Orthogonality Constraints Help Language-Model Training Beyond Sphere Optimization?}

\author{\name Kai Hu \email kaihu@cmu.edu \\
       \name Li Zhang \\
       \name Weichen Yu \\
       \name Matt Fredrikson \\
       \name Haoqi Hu \\
       \addr Carnegie Mellon University}

\maketitle

\begin{abstract}
A growing body of work reports benefits from orthogonality constraints in
fine-tuning large language models (LLMs). These results raise an open
question: do orthogonality constraints also help LLM pre-training? We
conduct a systematic empirical study of decoder-only LLM pre-training with
models ranging from 360M to 3.1B parameters, trained on nearly 200B tokens.
We examine fourteen variants of orthogonality training: some constrain the
weight matrix itself, while others use orthogonal factors to reparameterize it.
Because a strictly orthogonal square matrix loses half of its degrees of
freedom, we also study block-wise orthogonality, where every $d$ rows form a
row-orthonormal (Stiefel) matrix and the block size $d$ is a hyperparameter. Among
block sizes ranging from 1 to the shorter side of the matrix, no $d>1$ used
throughout the network reliably outperforms $d=1$, and large blocks are clearly
worse; the one placement that helps, symmetric positive-definite factors on the
query and key projections, loses its small gain as models grow. At $d=1$, the
constraint reduces to sphere optimization, with each row direction on its own
unit sphere, which by itself already improves on tuned Muon. The same two findings hold for a
diffusion transformer trained on ImageNet, and after fine-tuning, downstream
accuracy does not separate the parameterizations. Our conclusion is that, across the
realizations we tried, orthogonality between rows does \textbf{not} help language-model
pre-training beyond sphere optimization. Since this appears to conflict with
several prior results, we re-examine five prior methods in pre-training and fine-tuning
that report orthogonality improvements. In each case the reported gain is
attributable to numerical precision, training horizon, optimizer realization,
or per-row norm control rather than to the orthogonality the methods impose.
\end{abstract}


\section{Introduction}

Before the rise of large language models (LLMs), weight orthogonality was
used to keep networks with millions of parameters well conditioned and their
feature directions nonredundant~\citep{cisse2017parseval,huang2018own}.
For modern pretrained models, orthogonal fine-tuning methods report gains
from learning orthogonal transformations of pretrained
weights~\citep{qiu2023oft,liu2024boft,wu2026psoft} or from constraining
low-rank adapter factors to Stiefel manifolds~\citep{li2025stella}.
In LLM pre-training, orthogonality has so far mostly shaped the training
dynamics rather than the weights: the Muon
optimizer~\citep{jordan2024muon,liu2025muon} orthogonalizes the update to each
weight matrix, and POET~\citep{qiu2025poet} learns orthogonal transformations
of a fixed random weight matrix, preserving its singular values. Research
blogs~\citep{bernstein2025manifolds,su2025stiefel} derive optimizers that keep
the weights themselves on the Stiefel manifold, without large-scale
language-model experiments. \citet{yoshihara2026manifold} evaluates this
approach in GPT-2 models with up to 124M parameters and assigns different
orthogonality structures to attention and MLP matrices. Whether keeping the
weights themselves orthogonal helps LLM pre-training at scale remains open. We study this question systematically
with decoder-only models from 360M to 3.1B parameters trained on nearly 200B
tokens.

The most direct form of orthogonality training requires an entire weight
matrix \(W\in\mathbb{R}^{m\times n}\) with \(m\leq n\) to have orthonormal
rows, \(WW^\top=I_m\). This is a very strong regularization. It imposes
\(m(m+1)/2\) constraints in total, so a square orthonormal matrix loses about
half of its degrees of freedom, i.e., the number of effective parameters is
roughly halved. We explored two routes for recovering the lost degrees of
freedom.

The first route relaxes the orthogonality constraint on \(W\) to an orthogonal
reparameterization of \(W\). The simplest variant adds per-row magnitudes: the
model learns a row-orthonormal factor \(Q\) and a positive diagonal factor
\(D\), and the effective weight is \(W=DQ\). This reparameterization can
represent every matrix whose rows are orthogonal and nonzero, but it
recovers only \(m\) degrees of freedom. Even with these magnitudes,
full-matrix orthogonality substantially worsens validation loss, whether it is
enforced by our projected update or by manifold Muon~\citep{bernstein2025manifolds}
(Figure~\ref{fig:blocksize}; Appendix~\ref{app:gains}). To go one step further,
recall that every full-rank matrix admits a polar decomposition; we therefore
replace the diagonal factor \(D\) with a symmetric positive-definite (SPD)
matrix \(S\), so that \(W=SQ\). This reparameterization recovers all \(mn\)
degrees of freedom: \(S\) contributes \(m(m+1)/2\) and \(Q\) contributes
\(mn-m(m+1)/2\). A triangular factor (\(W=TQ\)) and a singular-value form
(\(W=U\Sigma V^\top\)) are equally expressive, but in training they recover
much less of the constraint penalty than SPD, and the singular-value form
recovers almost none.

The second route relaxes the constraint itself. An \(n\times n\) square matrix
is partitioned into \(n/d\) blocks of size \(d\times n\), and only each block
is required to have orthonormal rows, so the number of constraints falls to
\(n(d+1)/2\). The block size \(d\) becomes a single knob that ranges from
\(d=1\), which constrains only the row norms, to the shorter side of the
matrix. Increasing \(d\) adds exactly one thing: zero inner products between
rows in the same block. The two routes combine: each block is reparameterized
with per-row magnitudes, \(W_k=D_kQ_k\), or with an SPD factor,
\(W_k=S_kQ_k\). Our main comparison uses per-row magnitudes at every block
size, so increasing \(d\) changes only the orthogonality between rows. In the
special case \(d=1\), the SPD factor reduces to the per-row magnitude, and the
constraint reduces to a sphere constraint on each row with a learnable
magnitude.

Combining several ways to train orthonormal matrices with several ways to
recover the lost degrees of freedom, we test fourteen variants, together with three alternative updates for the
sphere control (Table~\ref{tab:variants}), in pre-training
Llama-style LLMs of three sizes, 360M, 1.3B, and 3.1B parameters, with token
budgets from \(2\times\) to \(16\times\) the Chinchilla-optimal budget.
No block size \(d>1\) used throughout the network reliably improves on
\(d=1\).
Small blocks are indistinguishable from \(d=1\), and the penalty grows
steadily with block size. SPD factors, which restore mixing between rows of a
block, recover much of this penalty at moderate block sizes but give no
consistent advantage, and the other realizations trail \(d=1\) as well.
The one placement that helps, SPD factors on the query and key projections,
gives a small gain in our smallest model that fades as models grow. At
\(d=1\) the orthogonality constraint reduces to sphere optimization: each row
direction lives on its own unit sphere with a learned magnitude, and no
relation between rows is imposed. We thus arrive, from the orthogonality side,
at the direction--magnitude geometry of recent row-norm
methods~\citep{lion2026muown,hubler2026muown,hagele2026decoupling}, and like
them we find that it improves on tuned Muon, as well as on MuonH, which fixes
the Frobenius norm of each matrix~\citep{wen2026hyperball}, at every scale we
trained. Across the realizations we tried, orthogonality between rows adds
nothing beyond sphere optimization. However, the implementation of sphere
optimization also matters: weight normalization~\citep{salimans2016weightnorm}
trained with Muon has the same expressiveness but recovers less than half of
the improvement over Muon.

This conclusion appears to conflict with several reports of orthogonality
gains, so we re-examine five of them in pre-training and fine-tuning. The
released implementations of POET and Pion store the AdamW baseline in
bfloat16; with fp32 storage, AdamW improves by 80 to 150 milli-loss, POET's
short-run advantage shrinks from 31 to 9 milli-loss and reverses with
longer training, and Pion trails AdamW by 95 milli-loss at 1.3B.
For module-wise constraints, projected diagonal-Gram training trails a
reparameterization that targets the same row geometry, and neither
diagonal-Gram nor Stiefel blocks improve on \(d=1\) in the module to which
they are assigned. In StelLA, replacing
Euclidean adapter factors with sphere-constrained ones gains 2.1 accuracy
points, and adding Stiefel constraints on top changes accuracy by
\(-0.09\). In PSOFT, replacing the orthogonal latent transformation with
independently normalized rows does not reduce accuracy. In each case, the
reported gain traces to numerical precision, training horizon, optimizer
realization, or per-row norm control rather than to orthogonality between
rows.

Our contributions are:
\begin{itemize}
    \item A block-wise orthogonality family whose block size moves from sphere
    optimization to the largest Stiefel blocks each matrix allows while holding row
    magnitudes fixed, so that its effect isolates orthogonality between rows.
    \item A pre-training study from 360M to 3.1B parameters showing that no
    block size above one used throughout the network improves on sphere
    optimization, that SPD factors on query and key projections give a small
    gain that vanishes as models grow, and that sphere optimization itself
    improves on tuned Muon; the same holds for a diffusion transformer on
    ImageNet, and fine-tuned models show no downstream advantage for any block
    size.
    \item Re-examinations of five orthogonality methods in pre-training and
    fine-tuning that trace each reported gain to a factor other than
    orthogonality between rows.
\end{itemize}

\section{Related Work}

\paragraph{Orthogonality in neural-network training.}
Orthogonal weights have been pursued through tight-frame, reparameterization,
penalty, and manifold methods~\citep{cisse2017parseval,huang2018own,bansal2018orthogonality,li2020cayley}, distinct from spectral normalization~\citep{miyato2018spectral}; classical treatments develop the underlying matrix-manifold algorithms~\citep{edelman1998geometry,absil2008optimization}. In the RNN study of \citet{vorontsov2017orthogonality}, soft orthogonality outperformed exact constraints. We compare hard block constraints with a norm-matched sphere control; soft orthogonality penalties are not evaluated here.

\paragraph{Geometry in parameter-efficient adaptation.}
Orthogonality can constrain transformations of pretrained weights, SVD-like or initialized LoRA factors, or manifold-optimized adapters~\citep{qiu2023oft,liu2024boft,wu2026psoft,zhang2023adalora,buyukakyuz2024olora,park2025stiefellora,li2025stella,bogachev2025riemannlora,zhang2026retractionfree}. These constructions act on different objects. Our adaptation study compares sphere and Stiefel constraints within StelLA's two-factor \(USV^\top\) parameterization.

\paragraph{Geometry-aware language-model pretraining.}
Muon orthogonalizes matrix-valued updates, while manifold Muon asks how to optimize a Stiefel-constrained weight~\citep{jordan2024muon,liu2025muon,wen2025fantastic,su2025stiefel,bernstein2025manifolds,solonko2026stiefel}; we ask whether that constraint improves over a sphere control with matched gains. Related methods control Frobenius or operator norms, normalize update directions, or impose module-wise Stiefel and DGram geometry~\citep{wen2026hyperball,xie2026sso,gu2026mano,yoshihara2026manifold,an2026demystifying,ren2026hypersphere,xiao2026hyperballfreelunch}. POET and Pion use orthogonal transformations but do not require orthogonal effective-weight rows~\citep{qiu2025poet,shi2026pion}.

\paragraph{Normalization and magnitude--direction separation.}
Weight normalization separates direction from magnitude~\citep{salimans2016weightnorm}. Muown is the closest prior parameterization to our \(d=1\) control~\citep{lion2026muown,hubler2026muown}; related work studies row or column gains, broader parameter normalization, and magnitude--direction decompositions in adaptation~\citep{hagele2026decoupling,velikanov2026learnable,loshchilov2024ngpt,liu2024dora,hu2022lora,kang2026nora}. Our \(d=1\) control confirms this norm-and-scale line; the new comparison begins when \(d\) increases while per-row gains remain fixed. Concurrent work traces the gains of fixed-norm training largely to its effective learning rate~\citep{yu2026airschedule,liu2026elrgoverns}.

Appendix~\ref{app:related} gives an extended discussion.

\section{Method}

We build every constrained variant on Muon. Section~\ref{sec:orthmuon}
describes a general recipe for training a matrix with orthonormal rows, and
Section~\ref{sec:recover} applies it to row blocks of Transformer weights and
describes how we return the degrees of freedom that the constraint removes.
Appendix~\ref{app:baselines} describes the other optimizers and controls we
compare.

\subsection{Optimizing row-orthonormal matrices with OrthMuon}
\label{sec:orthmuon}

OrthMuon separates the two jobs of a constrained optimizer: Muon chooses the
update direction, and the geometry of the constraint set keeps the parameter
feasible. For \(m\leq n\), the constraint set is the row-Stiefel manifold
\begin{equation}
    \operatorname{St}(m,n)
    =
    \left\{
        Q\in\mathbb{R}^{m\times n}: QQ^\top=I_m
    \right\},
\end{equation}
which is the orthogonal group when \(m=n\).

At each step, OrthMuon forms Muon's direction \(Z_t\) from the Euclidean
gradient, by Nesterov momentum followed by Newton--Schulz
orthogonalization~\citep{jordan2024muon,liu2025muon}. It then removes the
part of \(Z_t\) that would violate the constraint to first order, by
projecting onto the tangent space
\(T_Q=\{\Delta:\Delta Q^\top+Q\Delta^\top=0\}\),
\begin{equation}
    \Pi_{Q}(Z)=Z-\operatorname{sym}(ZQ^\top)Q,
    \qquad \operatorname{sym}(X)=\tfrac12(X+X^\top),
\end{equation}
and takes the step \(A_t=Q_t-\eta_t s\,\Pi_{Q_t}(Z_t)\), where \(s\) is
Muon's shape-dependent scale.

Because the step is tangent,
\(A_tA_t^\top=I_m+\eta_t^2s^2\,\Pi_{Q_t}(Z_t)\Pi_{Q_t}(Z_t)^\top\): the
iterate leaves the manifold only at second order. We therefore return to the
manifold with a polynomial instead of a matrix factorization. The nearest
point of \(\operatorname{St}(m,n)\) to \(A_t\) is its polar factor
\((I_m+E_t)^{-1/2}A_t\) with \(E_t=A_tA_t^\top-I_m\), and we replace the inverse
square root by its fourth-order Taylor polynomial,
\begin{equation}
    Q_{t+1}=p(E_t)\,A_t,
    \qquad
    p(E)=I_m-\tfrac12E+\tfrac38E^2-\tfrac5{16}E^3+\tfrac{35}{128}E^4.
\end{equation}
The retraction uses only matrix products, the primitive that Newton--Schulz
already relies on, with no QR, SVD, or eigendecomposition in the inner loop.
Proposition~\ref{prop:retraction} bounds its error and cost
(proof in Appendix~\ref{app:impl}).

\begin{proposition}[Polynomial retraction]
\label{prop:retraction}
Let \(Q\in\operatorname{St}(m,n)\), let \(\Delta\) be tangent at \(Q\) with
\(\|\Delta\|_2\leq1\), and let \(p_r\), \(r\geq1\), be the degree-\(r\) Taylor polynomial
of \((1+x)^{-1/2}\). Then \(E=(Q+\Delta)(Q+\Delta)^\top-I_m=\Delta\Delta^\top\),
and \(Q^+=p_r(E)(Q+\Delta)\) satisfies
\[
    \bigl\|Q^+Q^{+\top}-I_m\bigr\|_2\leq c_r\|\Delta\|_2^{2r+2},
    \qquad
    c_r=\binom{2r+2}{r+1}\Big/2^{2r+1},
\]
with equality to leading order as \(\Delta\to0\). In Horner form,
\(Q^+\) costs \(4m^2n+2(r-1)m^3\) matrix-multiply FLOPs.
\end{proposition}

With \(\Delta=-\eta_ts\,\Pi_{Q_t}(Z_t)\), the order \(r=4\) used here gives
\(c_4=63/128\) and an error of order \(\eta_t^{10}\). Raising the order from two
to four adds only two \(m\times m\) products. We therefore fix \(r=4\) rather
than choose the order from \(\|E_t\|_F\), which would force a GPU-to-CPU
synchronization at every step. A few exact polar corrections over training
remove accumulated rounding error.
At 1.3B on GB300 GPUs, \(d=1\) trains at 98\% of Muon's throughput
(Appendix~\ref{app:impl}); we hope this implementation serves as a practical
reference for orthogonality training at scale.

\newfloat{algorithm}{t}{loa}\floatname{algorithm}{Algorithm}
\begin{algorithm}[t]
\caption{One OrthMuon step for \(Q\in\operatorname{St}(m,n)\).
Appendix~\ref{app:impl} gives the blockwise version and implementation
details.}
\small
\begin{tabular}{@{}r@{\hspace{0.6em}}p{0.62\textwidth}@{\hspace{0.5em}}>{\itshape\raggedright\arraybackslash}p{0.27\textwidth}@{}}
\toprule
\multicolumn{3}{@{}p{0.97\textwidth}@{}}{\textbf{Input:} \(Q\in\operatorname{St}(m,n)\),
momentum \(B\), gradient \(G\), learning rate \(\eta\), scale \(s\).}\\
\midrule
1: & \(B\gets\mu B+G\), \(\;Z\gets\operatorname{NS}_5(G+\mu B)\) & Muon direction\\
2: & \(A\gets Q-\eta s\,\bigl(Z-\operatorname{sym}(ZQ^\top)Q\bigr)\) & tangent step\\
3: & \(Q\gets p(AA^\top-I_m)\,A\) & polynomial retraction\\
\bottomrule
\end{tabular}
\end{algorithm}

\subsection{Recovering degrees of freedom}
\label{sec:recover}

Let \(W\in\mathbb{R}^{m\times n}\) be a matrix-shaped Transformer weight. We
partition its rows into \(m/d\) contiguous blocks and keep the directional
factor of each block, \(Q_k\), on \(\operatorname{St}(d,n)\). The Muon
direction and scale are computed for the whole matrix; each block is then
projected and retracted independently as in Section~\ref{sec:orthmuon}. When
\(m\leq n\), \(d=m\) constrains the whole matrix.

\paragraph{Reparameterization.}
The first route forms each effective block from its orthonormal factor and a
magnitude factor. With per-row magnitudes,
\begin{equation}
    W_k=D_kQ_k,
    \qquad
    D_k=\operatorname{diag}(a_k),
    \qquad
    W_kW_k^\top=D_k^2,
\end{equation}
so the effective rows stay orthogonal while each keeps its own learned norm.
To allow mixing between rows, we replace \(D_k\) with an SPD factor,
\begin{equation}
    W_k=S_kQ_k,
    \qquad
    S_k=L_kL_k^\top,
    \qquad
    L_k=\operatorname{diag}(a_k)+\operatorname{tril}(M_k,-1).
\end{equation}
Now \(W_kW_k^\top=S_k^2\) is generally dense, so the effective rows are no
longer orthogonal, and by the polar decomposition this form can represent any
full-row-rank block. We initialize \(S_k=I_d\). The magnitude factors
\(a_k\) and \(L_k\) are trained with AdamW. Other left factors, including
triangular (\(W_k=T_kQ_k\)), singular-value (\(W_k=U_k\Sigma_kQ_k\) with
\(U_k\) orthogonal), and other symmetric forms, are equally expressive and are
defined in Appendix~\ref{app:baselines}.

\paragraph{Block size.}
The second route shrinks the set of constrained row pairs. With per-row
magnitudes at every \(d\), an \(m\)-row matrix has \(m(d-1)/2\) row pairs
constrained to be orthogonal, a fraction \((d-1)/(m-1)\) of all pairs, while
the number of row norms and the way they are optimized stay the same.
Increasing \(d\) therefore changes only the orthogonality between rows. At
\(d=1\), the directions form a product of unit spheres (the oblique
manifold~\citep{absil2006joint}), the SPD factor reduces to the per-row
magnitude, and the model has the same function class as an unconstrained row
with no constraint between rows. We call this case sphere optimization.

\section{Controlled Pretraining Results}
\label{sec:experiments}

\subsection{Experimental design and tuning}

We train decoder-only Transformers with 360M, 1.3B, and 3.1B parameters at
context length 4096. Muon-family runs scale decoupled weight decay by the Muon learning
rate~\citep{liu2025muon} and use cosine
decay to \(0.01\) times the peak learning rate. Tuned Muon, which shapes the
update spectrum without constraining the trained weight to be orthogonal,
serves as our principal unconstrained baseline. Our primary corpus is
C4~\citep{raffel2020exploring}.
Budgets are reported as multiples of 20 tokens per parameter, the representative
Chinchilla ratio~\citep{hoffmann2022chinchilla}. Unless stated otherwise, we
report final target-step cross-entropy on C4-val, a fixed evaluation set
drawn from the C4 validation split. Muon, MuonH, and the
\(d=1\) control receive separate main-learning-rate sweeps; unconstrained
baselines also tune weight decay. The AdamW-side learning rate is fixed at
0.0012 across the Muon-family arms. For the 360M \(d\)-ladder, a bracket over
\(\{0.001,0.002,0.004\}\) selects the common interior value \(0.002\) for
each \(d\), including \(d=32\). Main-text entries are means over seeds; each seed sets the initialization
and, except at 1.3B, \(7.77\times\), the order of the training data; Appendix~\ref{app:seeds} gives seed counts, per-seed losses, and paired
intervals, and Appendices~\ref{app:protocols}--\ref{app:optimizer-controls}
give additional grids. We report loss differences in milli-loss (\(10^{-3}\)
nats), each measured against a matched \(d=1\) run
(Appendix~\ref{app:seeds}).

\subsection{Block orthogonality provides no measured gain}

The headline comparison holds norm treatment and gain granularity fixed.
Every \(d=1,8,32\) model retains one learned gain per row; increasing \(d\)
makes rows within each block orthogonal to each other. At \(2\times\),
\(4\times\), and \(7.69\times\), \(d=8\) changes mean loss by only 0.8--1.1
milli-loss, whereas \(d=32\) increases it by 3.7--5.2 milli-loss
(Table~\ref{tab:main}; Appendix~\ref{app:d-ladder} for \(4\times\)).
Because per-row blocks with \(d>1\) also remove directions within each block,
this ladder measures the constraint together with its cost in expressivity;
the matrix-gain controls of Section~\ref{sec:spd-results} restore that
expressivity and isolate the geometry.
Every block size also uses unit-norm rows and the same gains, so the
effective learning rate of the directions, which recent work identifies as a
main driver of fixed-norm
training~\citep{vanlaarhoven2017l2,yu2026airschedule,liu2026elrgoverns}, is
the same across the ladder.
Training to \(16\times\) narrows the \(d=32\) gap to 2.4 milli-loss, which
still corresponds to about 5\% more training steps
(Appendix~\ref{app:overtrain}).

\begin{table}[t]
    \centering
    \caption{Final C4 validation loss (lower is better): the \(d=1\)
    control beats Muon and MuonH, while \(d=8,32\) do not improve on
    it in the matched per-row-gain ladder.}
    \label{tab:main}
    \small
    \setlength{\tabcolsep}{5pt}
    \begin{tabular}{lcccccc}
        \toprule
        & \multicolumn{4}{c}{Trained on C4} & \multicolumn{2}{c}{Trained on FPO-mix} \\
        \cmidrule(lr){2-5}\cmidrule(lr){6-7}
        Method & \shortstack{360M\\\(2\times\)} & \shortstack{360M\\\(7.69\times\)} & \shortstack{1.3B\\\(2\times\)} & \shortstack{3.1B\\\(2\times\)} & \shortstack{360M\\\(2\times\)} & \shortstack{1.3B\\\(2\times\)} \\
        \midrule
        Muon & 2.475 & 2.372 & 2.197 & 2.036 & 2.682 & 2.413 \\
        MuonH (fixed matrix norm) & 2.469 & 2.362 & 2.191 & 2.031 & 2.677 & 2.407 \\
        OrthMuon \(d=1\) (sphere + gain) & \textbf{2.462} & \textbf{2.356} & \textbf{2.184} & \textbf{2.024} & \textbf{2.671} & \textbf{2.400} \\
        \midrule
        OrthMuon \(d=8\) & 2.463 & 2.358 & -- & -- & -- & -- \\
        OrthMuon \(d=32\) & 2.467 & 2.360 & 2.186 & 2.025 & 2.675 & -- \\
        \bottomrule
    \end{tabular}
\end{table}

\subsection{Expressivity and optimizer controls for \(d>1\)}
\label{sec:spd-results}

\begin{figure}[t]
    \centering
    \includegraphics[width=0.62\linewidth]{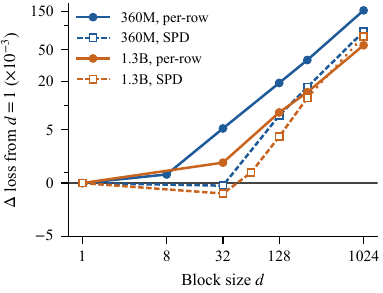}
    \caption{The C4 loss penalty grows with block size under
    per-row and SPD gains (\(2\times\); symmetric-log axis; each point
    uses its matched \(d=1\) reference; the dotted segment marks a
    configuration that is unstable at its selected learning rate,
    Appendix~\ref{app:gains}).}
    \label{fig:blocksize}
\end{figure}

We tested diagonal and matrix-valued gains together with projected
diagonal-Gram (DGram) training, which keeps \(WW^\top\) diagonal by projection
while leaving row norms free. SPD uses \(S=LL^\top\), XXT uses
\(S=XX^\top\), and XXT+diagonal uses
\(S=\operatorname{diag}(a)+XX^\top\). At 360M, SPD and XXT close the 6.5 milli-loss deficit of one scalar gain
per block, to within 0.3 milli-loss of \(d=1\) (Table~\ref{tab:dgt1-main}),
consistent with restoring within-block degrees of freedom, including row
mixing. Trained \(d=1\) rows
are about as decorrelated as random directions: within 32-row windows, their
mean absolute cosine is 0.030 (0.047 in query and key matrices), against 0.025
for random directions in \(\mathbb{R}^{1024}\) and 0.032 and 0.034 for tuned Muon and
MuonH. Per-row \(d=32\)
sets the within-block value to zero, and the SPD gain raises it to 0.052
(Appendix~\ref{app:row-geometry}). Expressivity alone does
not decide this: equally expressive triangular and singular-value factors
recover much less (Table~\ref{tab:gain-param}). Larger blocks widen
the gap (Figure~\ref{fig:blocksize}): at \(d\geq128\), every tested per-row and
SPD configuration trails \(d=1\) at both 360M and 1.3B, by more at each
larger \(d\).

\begin{table}[t]
    \centering
    \small
    \caption{Change in final C4 loss from \(d=1\) at \(d=32\), \(2\times\)
    (\(\times10^{-3}\); negative is better; matched \(d=1\)
    references): full-matrix gains close the 360M gap, but no
    realization reliably beats \(d=1\).}
    \label{tab:dgt1-main}
    \begin{tabular}{lccc}
        \toprule
        Realization at \(d=32\) & 360M & 1.3B & 3.1B \\
        \midrule
        Per-row gain & +5.2 & +1.9 & +0.8 \\
        Per-block gain & +6.5 & -- & -- \\
        SPD gain \(LL^\top\) & \(-0.2\) & \(-1.0\) & \(-0.3\) \\
        XXT gain & +0.2 & -- & -- \\
        Projected DGram & +18.8 & -- & -- \\
        Manifold Muon & +35.5 & +27.6 & -- \\
        \bottomrule
    \end{tabular}
\end{table}

Matrix-valued gains, which allow nonorthogonal effective rows, show no
consistent advantage over \(d=1\) when every matrix uses them. At \(d=32\),
SPD differs from \(d=1\) by \(-0.2\) milli-loss at 360M, \(-1.0\) at 1.3B,
and \(-0.3\) at 3.1B, and by \(-0.4\) and \(-0.7\) on the FPO mixture at
\(2\times\) and \(4\times\). These comparisons train every gain at the shared
default learning rate. Raising the gain learning rate of \(d=1\) to 0.0048
makes SPD 0.6 milli-loss worse than \(d=1\) at 1.3B, whereas SPD's own factor
learning rate is already best at the default (Appendix~\ref{app:gain-lr}). At
1.3B, \(d=64\) SPD trails \(d=1\) by 1.0 milli-loss.

Manifold Muon~\citep{bernstein2025manifolds,su2025stiefel} solves the
constrained spectral-norm steepest-descent problem directly but trails
\(d=1\) at every tested block size, budget, and scale: by 26--142 milli-loss
at 360M and by 23--63 milli-loss at 1.3B (Appendix~\ref{app:gains}). It
improves on projected OrthMuon only at the largest tested block size at 360M
and trails it at every tested block size at 1.3B, so the negative result is
not specific to the projected update.

\subsection{Row mixing in queries and keys helps only small models}
\label{sec:qk-spd}

One placement of SPD factors does improve on \(d=1\), but the gain fades as
models grow. We give the query and key projections SPD factors with block
size 64 and keep every other matrix at \(d=1\). The reference is \(d=1\)
throughout with its gain learning rate raised to 0.0048, which improves on
the default by 0.6--1.8 milli-loss at 360M, 1.5 at 1.3B, and 0.3 at 3.1B
(Appendix~\ref{app:gain-lr}). Against this reference, SPD factors on queries
and keys lower the loss by 4.6 milli-loss at 360M, \(2\times\). Queries and
keys contribute about equally, whereas SPD factors on value, output, or MLP
projections do not help (Table~\ref{tab:qk-spd}). The gain shrinks with
training length, to 3.3 milli-loss at \(4\times\) and 2.7 at \(7.69\times\),
and with model size, to 1.1 at 1.3B and 0.7 at 3.1B
(Appendix~\ref{app:qk-spd}).

Larger blocks reverse the effect. At 3.1B, whose heads have 128 rows,
\(d=128\) SPD factors on queries and keys are 0.7 milli-loss worse than the
reference. Unlike a per-row gain, an SPD factor does not bound the norm of
a query or key block: with \(d=128\), a few heads reach attention logits of
300 to 5{,}000, against at most 222 for \(d=1\)
(Appendix~\ref{app:qk-spd}).

\begin{table}[t]
    \centering
    \caption{Change in final C4 loss from \(d=1\) when one projection type
    uses SPD factors with block size 64 (\(\times10^{-3}\); negative is
    better): only queries and keys benefit, and less at 1.3B.}
    \label{tab:qk-spd}
    \small
    \begin{tabular}{lccc}
        \toprule
        SPD factor on & 360M, \(2\times\) & 360M, \(4\times\) & 1.3B, \(2\times\) \\
        \midrule
        Query and key & \(\mathbf{-4.6}\) & \(\mathbf{-3.3}\) & \(\mathbf{-1.1}\) \\
        Query only & \(-2.2\) & \(-2.2\) & \(\mathbf{-1.1}\) \\
        Key only & \(-2.3\) & \(-2.0\) & \(-0.4\) \\
        Value only & \(+1.0\) & \(+0.4\) & \(+0.9\) \\
        Output only & \(+0.1\) & \(+0.7\) & \(+1.3\) \\
        MLP gate only & \(+1.4\) & \(+1.3\) & \(+1.2\) \\
        MLP up only & \(+0.3\) & \(+1.2\) & \(+1.3\) \\
        MLP down only & \(+2.1\) & \(+1.0\) & \(+1.5\) \\
        \bottomrule
    \end{tabular}
\end{table}

\subsection{The \(d=1\) control is strong but optimizer-dependent}
\label{sec:decomposition}

The \(d=1\) control improves over tuned Muon at the shared AdamW-side
learning rate (Section~\ref{sec:experiments}), confirming the positive
norm-and-scale results reported by recent work~\citep{lion2026muown,hubler2026muown,hagele2026decoupling}.
Across three budgets, it lowers validation loss by 13.1--17.7 milli-loss;
the unit-row constraint without gain and the learned row magnitudes each
contribute about half (Table~\ref{tab:decomp-main}). MuonH, which controls
one Frobenius norm per matrix, is within 0.6 milli-loss of the no-gain
\(d=1\) model at \(2\times\) and \(4\times\) and 2.7 milli-loss better at
\(7.69\times\). Its
deficit to \(d=1\) lies mainly in attention: 7.5 milli-loss there versus 1.4
in MLP matrices (Appendix~\ref{app:module-swap}).

\begin{table}[t]
    \centering
    \caption{Improvement over tuned Muon at 360M (\(\times10^{-3}\); higher is
    better). A unit-row directional constraint and learned row gains each
    supply about half of the \(d=1\) gain; normalizing rows only in the forward pass recovers
    less than half.}
    \label{tab:decomp-main}
    \small
    \begin{tabular}{lccc}
        \toprule
        Method & \(2\times\) & \(4\times\) & \(7.69\times\) \\
        \midrule
        Forward-normalized Muon & 5.5 & -- & -- \\
        MuonH (fixed matrix norm) & 6.7 & 10.8 & 9.8 \\
        \(d=1\), no gain & 7.2 & 10.1 & 7.1 \\
        \(d=1\), row gain (total) & \textbf{13.1} & \textbf{17.7} & \textbf{15.7} \\
        \bottomrule
    \end{tabular}
\end{table}

The function class alone does not explain this gain. Applying Muon to an
unconstrained latent weight and normalizing rows only in the forward pass
reaches 2.4697. It recovers 5.5 of the 13.1 milli-loss
improvement over Muon and remains 7.6 milli-loss behind explicit
sphere-coordinate optimization. The coordinate system and update dynamics therefore matter
alongside norm and scale.
Concurrent work attributes most of the advantage of fixed-norm training over
weight decay to its effective-learning-rate
trajectory~\citep{yu2026airschedule,liu2026elrgoverns}; part of the gain of
\(d=1\) over Muon may arise the same way, whereas the learned row gains, which
supply about half of it, are part of the parameterization.
Alternative gain coordinates, direct row normalization, and tangent or
transported momentum remain within 1.6 milli-loss of the explicit \(d=1\)
result (Appendix~\ref{app:optimizer-controls}).

\subsection{Scope across scale, data, modality, and fine-tuning}

Changing the training mixture preserves the matched diagonal-gain ordering.
With the FPO-mix training mixture---92.57\% DCLM, 6.09\% StarCoder, and 1.34\%
ProofPile~\citep{wen2025fantastic}---and C4-val evaluation,
\(d=1\) again leads MuonH and Muon at 360M and 1.3B, and per-row-gain
\(d=32\) at 360M (Table~\ref{tab:main}; Appendix~\ref{app:fpo}).
On held-out web text, C4-trained 360M checkpoints keep the C4 ordering at
about its C4 magnitude: on the FPO mixture and DCLM, per-row \(d=8\) and
\(d=32\) match \(d=1\), Muon and MuonH trail it by 7 to 14 milli-loss, and
AdamW by about 90. On StarCoder, ProofPile, and OLM-arXiv, loss varies by
60--100 milli-loss between training runs of the same configuration, and no
ordering among block sizes or baselines emerges (Appendix~\ref{app:ood}).

The strength of the \(d=1\) control also extends beyond 360M. At 1.3B, it
improves over Muon by 12.8, 13.1, and 12.0 milli-loss at \(2\times\),
\(4\times\), and \(7.77\times\), and by 12.0 at 3.1B;
Appendix~\ref{app:scaling} reports the full scaling table. The per-row \(d=32\) penalty shrinks with scale
(Table~\ref{tab:dgt1-main}), as does the query--key SPD gain, so both effects
of row structure approach zero in the tested range; at 360M the \(d=32\)
penalty corresponds to about 5\% more training steps at every budget up to
\(16\times\) (Appendix~\ref{app:overtrain}).

Sphere optimization also improves image generation. In a 130M-parameter
diffusion transformer (SiT-B/2) trained on ImageNet at \(256\times256\) with
query--key normalization, \(d=1\) lowers FID-50K from 25.50 to 24.60 with an
Adam inner optimizer and from 21.55 to 20.81 with Muon, and lowers validation
loss in both families. Query and key blocks of 64 rows do not improve on
\(d=1\), and an SPD factor on those blocks ties it (Appendix~\ref{app:dit}).
Without query--key normalization, AdamW and \(d=1\) occasionally diverge late
in training, whereas scalar-gain \(d>1\) blocks, which bound the query--key
product, never do. Query--key normalization removes these divergences and gives
lower FID than \(d>1\) without it, so the stability benefit of \(d>1\) traces
to logit control rather than to orthogonality between rows.

Fine-tuning does not separate the parameterizations either. After full
fine-tuning of the 3.1B models on Commonsense170K, with the merged weights
trained by one AdamW recipe, eight-task accuracy differs from the matched
\(d=1\) model by at most 0.4 points for every \(d>1\) and query--key
configuration, well within the variation between fine-tuning runs. Muon and
MuonH trail \(d=1\) by 0.7 and 1.1 points. Fine-tuning the merged weights with
AdamW raises every model by 1.5 to 3.1 points over continuing with its
pretraining optimizer (Appendix~\ref{app:sft}).

\section{Complementary Attribution Studies}

We next revisit orthogonality-associated methods in pretraining and adaptation.
POET and Pion do not constrain the rows of the effective weight; we include
them as audits of orthogonality-based methods, whose comparisons share the
baseline issue examined here.
The cases address different attribution questions: POET and Pion test numerical
precision and training horizon, DGram tests optimizer realization for a shared
constraint set, StelLA supplies a second matched sphere-to-Stiefel comparison,
and PSOFT tests the parameterization of a latent transformation. Only the pretraining \(d\)-ladder and the StelLA
comparison are direct sphere-to-Stiefel comparisons; the other cases address
different attribution questions.

Table~\ref{tab:audits} summarizes the four studies.

\begin{table}[t]
    \centering
    \caption{Controlled comparisons separate orthogonality from precision,
    training horizon, optimizer realization, and norm control.}
    \label{tab:audits}
    \setlength{\tabcolsep}{4pt}
    \renewcommand{\arraystretch}{1.08}
    \begin{tabular}{@{}>{\raggedright\arraybackslash}p{0.18\textwidth}
                        >{\raggedright\arraybackslash}p{0.29\textwidth}
                        >{\raggedright\arraybackslash}p{0.47\textwidth}@{}}
        \toprule
        Study & Comparison & Finding \\
        \midrule
        POET, Pion & fp32 AdamW state; longer training & POET: 31 to 9
        milli-loss with fp32 state; behind AdamW at 6B and 30.1B tokens. Pion:
        behind bfloat16 and fp32 AdamW at 1.3B. \\
        Different Layers & DGram projection versus reparameterization;
        module assignment &
        Projection trails Muon, \(d=1\), and the \(d=32\)
        reparameterization. Neither DGram in MLP nor per-row \(d=32\) in
        attention improves on \(d=1\). \\
        StelLA & Sphere-constrained factor columns &
        Sphere gains 2.1 points over Euclidean; Stiefel then changes accuracy
        by \(-0.09\). \\
        PSOFT & Row-sphere transform in the same selected subspace &
        Row-sphere scores 0.5 points above the implemented transform at
        validation-selected rates; the ordering reverses with learning rate. \\
        \bottomrule
    \end{tabular}
\end{table}

\begin{table}[t]
    \centering
    \small
    \caption{fp32 storage makes AdamW much stronger and shrinks or
    reverses each method's advantage. Entries
    are final losses; \(\Delta\) is method minus AdamW (\(\times10^{-3}\);
    negative favors the method). POET uses the same storage as AdamW;
    Pion runs once at its released precision.}
    \label{tab:precision}
    \setlength{\tabcolsep}{6pt}
    \begin{tabular}{lccc}
        \multicolumn{4}{l}{(a) POET, 130M, 0.2B tokens} \\
        \toprule
        Storage & AdamW & POET & \(\Delta\) \\
        \midrule
        bfloat16 & 3.943 & 3.912 & \(-31.2\) \\
        fp32 & 3.793 & 3.785 & \(-8.7\) \\
        \bottomrule
        \addlinespace[8pt]
        \multicolumn{4}{l}{(b) Pion, 1.3B, 51B tokens, context 4096} \\
        \toprule
        AdamW storage & AdamW & Pion & \(\Delta\) \\
        \midrule
        bfloat16 & 2.346 & 2.361 & \(+15.0\) \\
        fp32 & 2.265 & 2.361 & \(+95.4\) \\
        \bottomrule
    \end{tabular}
\end{table}

\begin{figure}[t]
    \centering
    \includegraphics[width=0.62\linewidth]{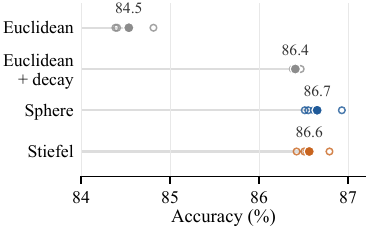}
    \caption{StelLA accuracy: Sphere reproduces the gain over
    Euclidean factors, and Stiefel adds nothing (filled markers are
    means, hollow markers individual seeds).}
    \label{fig:stella}
\end{figure}

\subsection{Pretraining studies}

\paragraph{POET and Pion: precision and training horizon.}
POET learns orthogonal transformations of a fixed random base matrix and
preserves its initialized singular-value spectrum~\citep{qiu2025poet}. Pion,
from the same group, turns this construction into an optimizer that updates
each weight through left and right orthogonal
transformations~\citep{shi2026pion}. Both released implementations store the
AdamW baseline's parameters and optimizer moments in bfloat16, and in both
cases fp32 storage makes AdamW far stronger (Table~\ref{tab:precision}). In
POET's harness at 197M tokens, with independently swept learning rates,
bfloat16 raises AdamW loss by 149 milli-loss and POET loss by 127, because
POET's factors start at zero or identity and its moments are periodically
reset. POET's advantage shrinks from 31.2 to 8.7 milli-loss, and it does not
survive longer training: in a separate 130M implementation, POET leads tuned
AdamW by 370 and 61 milli-loss at 0.59B and 2B tokens but trails by 8 and 19
milli-loss at 6B and 30.1B. Pion reports a 35-milli-loss advantage over
bfloat16 AdamW for a 1.3B model trained on 54B C4 tokens at context length
256, while keeping its own Lie-algebra momentum in fp32. At context length
4096, with the paper's update settings and a Pion learning-rate sweep, Pion
trails bfloat16 AdamW by 15.0 milli-loss and fp32 AdamW by 95.4
(Appendix~\ref{app:pion}). In both cases, storing the baseline in bfloat16
costs AdamW more than the reported advantage of the orthogonal method.

\paragraph{Different Layers, Different Manifolds: constraint versus optimizer.}
\citet{yoshihara2026manifold} reports that assigning Stiefel constraints to
attention and DGram constraints to MLP matrices lowers GPT-2 validation loss
from 3.3855 without constraints and 3.3679 under all-Stiefel training to
3.3544; their Stiefel arm follows the manifold-Muon construction of
\citet{bernstein2025manifolds}, and the result combines a module assignment
with a projected-gradient optimizer. Our comparison is not a reproduction of
that setting: it uses 360M models, blockwise \(d=32\) DGram instead of
whole-matrix constraints, and our training recipe. Our controlled comparison fixes a
blockwise \(d=32\) DGram geometry and varies its optimizer realization.
Projected DGram keeps \(WW^\top\) diagonal, whereas
\(W=\operatorname{diag}(a)Q\) targets the same row geometry but assigns
momentum, adaptive scaling, and regularization to different variables.
Independent LR--WD tuning lowers projected DGram from 2.544 to 2.481; it
remains behind tuned Muon (2.475) and \(d=1\) (2.462), and
behind the per-row \(d=32\) reparameterization of the matched ladder (2.467).
To test the module assignment, we keep one module type under \(d=1\) and
replace the other (Appendix~\ref{app:module-swap}). Relative to training every
matrix with \(d=1\), DGram raises loss by 12.6 milli-loss in MLP matrices and
10.3 in attention matrices, and per-row \(d=32\) raises it by 2.6 and 3.2.
Neither constraint improves on \(d=1\) in the module to which
\citet{yoshihara2026manifold} assign it, and the \(d=32\) penalty arises in
both module types.

\subsection{Adaptation studies}

\paragraph{StelLA: norm control.}
StelLA constrains both low-rank factors in Equation~\ref{eq:stella} to
Stiefel manifolds~\citep{li2025stella}. In Llama-3-8B commonsense
adaptation~\citep{grattafiori2024llama}, we insert a product-of-spheres control
between Euclidean and Stiefel factors. All runs use StelLA's default
\(S_0=I\). Sphere improves over Euclidean by 2.12 accuracy points, whereas
Stiefel changes Sphere by \(-0.09\) points (Figure~\ref{fig:stella}). All
arms use StelLA's default learning rate \(5\times10^{-4}\). In a check over
\(\{10^{-4},5\times10^{-4},10^{-3}\}\), this is the best tested value for
Sphere and Stiefel, while the best tested Euclidean rate, \(10^{-3}\), lies at
the grid edge and remains 1.87 points below Sphere
(Appendix~\ref{app:stella}).

The Sphere gain is large, while adding row orthogonality does not
improve accuracy. Factor-wise decay (U/V weight decay 1.0, the largest tested
value) comes within 0.25 points of Sphere and recovers about 88\% of the
Euclidean-to-Sphere gain. In an update-rule ablation, a direct-polar
Stiefel update improves over the canonical update by about 0.13 points
(Appendix~\ref{app:stella}).

\paragraph{PSOFT: orthogonality within principal-subspace adaptation.}
PSOFT learns an orthogonal transformation restricted to a principal subspace
of the pretrained weight~\citep{wu2026psoft}. Our controlled reconstruction
holds fixed the selected subspace, rank, target modules, effective
initialization, learned magnitude vectors, and training protocol. It compares
PSOFT's implemented Cayley/Neumann latent transformation with a dense latent
transformation whose rows are optimized independently on spheres. Because the
coordinate system and optimizer realization also change, this comparison tests the implemented orthogonal parameterization as a whole.
At learning rates selected by validation loss, the row-sphere transform reaches
81.07 average accuracy, compared with 80.55 for PSOFT's implementation and
80.95 for an unconstrained dense transform (Appendix~\ref{app:psoft}).
Removing orthogonality between rows does not reduce accuracy, and the sphere
constraint adds about 0.1 points over the dense transform at the selected
rates and none at \(8\times10^{-4}\). The ordering of
PSOFT and row-sphere depends on the learning rate: PSOFT leads by 0.19 points
at \(4\times10^{-4}\) and trails by 0.64 at \(8\times10^{-4}\). The implemented
five-term Neumann transform is also not exactly orthogonal; its singular values
range from 0.86 to 1.00.

\section{Discussion and Limitations}

\paragraph{Stronger block constraints and row-mixing freedom.}
A monotone conditioning or diversity account predicts that decorrelating more
row pairs should help, or at least not hurt. Instead, \(d=8\) has no resolved
effect, \(d=32\) is worse in mean loss at all three 360M budgets, and, at
360M, matrix gains that restore within-block degrees of freedom, including
row mixing, close the 6.5 milli-loss penalty of one scalar gain per block
(Section~\ref{sec:spd-results}).
In trained 360M checkpoints, per-row \(d=32\) gives 32-row blocks a higher
median stable rank than \(d=1\) (25.3 versus 21.6) and a lower median condition
number (1.30 versus 1.54), yet it has higher loss
(Appendix~\ref{app:row-geometry}).

\paragraph{Optimizer realization matters alongside the constraint.}
The \(d=1\) gain agrees with recent norm-and-scale
methods~\citep{lion2026muown,hubler2026muown,hagele2026decoupling}, yet
forward-normalized Muon, with the same function class, captures less than
half of it (Section~\ref{sec:decomposition}). Projected DGram and manifold
Muon likewise show that a nominal constraint does not determine finite-budget
performance independently of its optimizer. Norm and magnitude variables
should be evaluated together with the state space in which momentum and
preconditioning act.

\paragraph{Orthogonality claims are object-specific.}
Muon orthogonalizes an update, POET applies orthogonal transformations of a fixed random base matrix, OrthMuon constrains a directional factor whose effective
rows remain orthogonal only under diagonal gains, and StelLA constrains the
bases of a low-rank update. These interventions have different invariances,
function classes, and induced optimizers, so evidence for one does not
transfer to another; Muon's success, for example, does not predict that
orthogonal weights train better. We therefore suggest nested controls: an
unconstrained baseline for the full recipe, a norm-matched sphere control for
pairwise angles, matched gain granularity, an expressive matrix-gain control,
and optimizer-specific tuning. The target metric and numerical policy should also be explicit:
POET shows that an early-training advantage need not survive a completed
schedule, and POET and Pion show that a full-bfloat16 AdamW baseline can
trail the same baseline in fp32 by more than the reported gain. In both the pretraining ladder and the StelLA comparison, the
comparison against the unconstrained baseline is positive, while the increment
from row orthogonality is small or negative.

\paragraph{Limitations.}
Appendix~\ref{app:limitations} discusses the scope of our experiments.

\section{Conclusion}

We tested the incremental value of row orthogonality while holding per-row
norm treatment and gain granularity fixed. In the matched diagonal-gain
ladder, \(d=8\) has no resolvable effect relative to sphere optimization, and
\(d=32\) worsens mean validation loss at 360M; the per-row gap is smaller at
1.3B and 3.1B and grows with block size. StelLA shows the same
attribution in adaptation: sphere optimization supplies the gain over
Euclidean factors, while adding Stiefel geometry provides no gain.

Matrix-valued gains recover much of the large-block penalty but, used
throughout the network, do not beat \(d=1\). SPD factors on query and key
projections give a small gain at 360M that shrinks with training length and
fades as models grow. Manifold Muon trails \(d=1\) at every block size,
budget, and scale. Beyond this small-model exception, no \(d>1\) realization
provides a consistent advantage.

Orthogonal updates, latent factors, and effective weights are distinct
interventions. Attributing their benefits to orthogonality calls for sphere controls with
matched magnitude freedom. In our settings, the direct evidence favors sphere optimization over hard
block orthogonality.

\paragraph{Reproducibility statement.}
Appendix~\ref{app:impl} specifies the OrthMuon update, parameter grouping, and
available implementation details; the remaining appendices report tuning grids,
seed counts, per-seed results, and paired intervals.

\paragraph{AI use statement.}
The core implementation of OrthMuon was written by the authors without
generative-AI assistance. After the initial exploratory phase was completed,
generative AI tools were used to help construct a fully refactored and frozen
codebase for the large-scale experiments. Specifically, Claude Opus 4.8 was
used to develop the implementation plan, GPT-5.6 Sol was used to write the
refactored code, and Claude Opus 5.0 was used to review the resulting code.
The refactored codebase was validated by reproducing a subset of experiments
and confirming that the results matched those obtained with the human-written
code. The diffusion-transformer pipeline was written with Claude Opus 5.5 and
reviewed with GPT-6.1 Sol; the fine-tuning pipeline was written with GPT-6.1 Sol
and reviewed with Claude Opus 5.5. The authors reviewed both, and each was
checked for exactness, for example by matching step-0 losses of converted
checkpoints and by reproducing the upstream diffusion recipe.

Generative AI tools were also used to suggest additional ablation studies.
The authors made the final decisions regarding which ablations to pursue and
retained responsibility for their execution and interpretation. During
manuscript preparation, GPT-5.6 Sol and Claude Code were used for grammar and
language checking, cross-checking the consistency of reported experimental
results, and adversarial review aimed at identifying potential errors,
inconsistencies, or unsupported claims.

The authors reviewed all AI-assisted outputs and take full responsibility for
the code, experimental results, claims, and final content of this work.

\bibliography{references}

\clearpage
\appendix
\section{Extended Related Work}
\label{app:related}

\paragraph{Orthogonality in neural-network training.}
Orthogonal constraints have long been used to control signal propagation, conditioning, and robustness. Parseval Networks maintain linear operators as approximate tight frames, while Orthogonal Weight Normalization enforces rectangular weight orthogonality through a differentiable reparameterization~\citep{cisse2017parseval,huang2018own}. Softer approaches penalize deviations from orthogonality~\citep{bansal2018orthogonality}, whereas Cayley-transform methods optimize parameters directly on Stiefel manifolds~\citep{li2020cayley}. These objectives differ from spectral normalization, which controls the largest singular value without requiring weight vectors to be orthogonal to each other~\citep{miyato2018spectral}. We revisit explicit weight orthogonality in Transformer training while separating it from per-vector norm control.
Classical treatments develop optimization algorithms for Stiefel and related
matrix manifolds~\citep{edelman1998geometry,absil2008optimization}. In recurrent
networks, soft orthogonality constraints outperformed exact constraints in the
study of \citet{vorontsov2017orthogonality}; our experiments compare hard
block constraints with a norm-matched sphere control; soft orthogonality
penalties are not evaluated here.

\paragraph{Geometry in parameter-efficient adaptation.}
Orthogonality enters parameter-efficient adaptation through several distinct constructions. OFT learns an orthogonal transformation of frozen pretrained weights; BOFT represents this transformation with butterfly factors; and PSOFT restricts it to a principal subspace~\citep{qiu2023oft,liu2024boft,wu2026psoft}. These methods apply orthogonality to transformations of pretrained geometry rather than directly to effective weight rows. Within low-rank adaptation, AdaLoRA uses softly orthogonal SVD-like factors for rank allocation, and OLoRA uses an orthonormal initialization without maintaining the constraint~\citep{zhang2023adalora,buyukakyuz2024olora}. Recent methods optimize one or both LoRA factors on Stiefel manifolds~\citep{park2025stiefellora,li2025stella}, while RiemannLoRA optimizes the adapter on the fixed-rank matrix manifold to remove factorization ambiguity~\citep{bogachev2025riemannlora}. Manifold-LoRA compares Stiefel and product-of-spheres constraints on a single LoRA basis~\citep{zhang2026retractionfree}. Our adaptation study makes this comparison within StelLA's two-factor \(USV^\top\) parameterization and separates norm control from update design.

\paragraph{Geometry-aware language-model pretraining.}
Muon orthogonalizes matrix-valued updates~\citep{jordan2024muon,liu2025muon,wen2025fantastic}. Manifold Muon solves tangent-space spectral-norm steepest descent using fixed-point, dual-ascent, or exact closed-form solvers~\citep{su2025stiefel,bernstein2025manifolds,solonko2026stiefel}. It asks how to optimize a Stiefel-constrained weight; we ask whether that constraint improves over a sphere-constrained control with matched gain semantics. Hyperball fixes matrix Frobenius norms, while spectral-sphere optimization controls operator norms~\citep{wen2026hyperball,xie2026sso}. Mano instead constructs tangent momentum from normalized weights and alternates row- and column-wise normalization of the update while keeping the weights in Euclidean space~\citep{gu2026mano}. POET learns orthogonal transformations of a fixed random base matrix and preserves its initialized singular-value spectrum; Pion turns this isospectral structure into a direct optimizer based on left and right orthogonal transformations~\citep{qiu2025poet,shi2026pion}. Neither method requires orthogonal effective-weight rows. Module-wise Stiefel and DGram constraints have also been studied in Transformer pretraining~\citep{yoshihara2026manifold}. \citet{an2026demystifying} compare Frobenius and spectral spheres with input- and output-oblique constraints, including their interactions with RMSNorm and weight decay. Our study holds per-row magnitude freedom fixed while directly testing the addition of orthogonality between rows.

\paragraph{Normalization and magnitude--direction separation.}
Weight normalization separates a vector into a normalized direction and a learned magnitude~\citep{salimans2016weightnorm}. Muown is the closest prior parameterization and optimizer family to our \(d=1\) control, combining row-norm control with separately optimized magnitudes; AngularMuown relates this family to normalized directional updates~\citep{lion2026muown,hubler2026muown}. Other recent work combines fixed-norm directions with row or column gains, while Learnable Multipliers studies the benefit of freeing scalar, row, and column scales without normalizing directions~\citep{hagele2026decoupling,velikanov2026learnable}. Related language-model methods normalize broader collections of parameters or representations, including nGPT~\citep{loshchilov2024ngpt}, while DoRA applies magnitude--direction decomposition to low-rank adaptation~\citep{liu2024dora}. Normalized Low-Rank Adaptation (NoRA) normalizes LoRA's down-projection and also studies normalization at initialization only~\citep{hu2022lora,kang2026nora}. Our \(d=1\) control confirms this norm-and-scale line of work; the new comparison begins when \(d\) increases while per-row gains remain fixed.

\section{Experimental Protocols}
\label{app:protocols}

Pretraining experiments apply decoupled weight decay to Muon-updated matrices
scaled by the Muon learning rate, \(W\gets W-\eta\lambda W\)~\citep{liu2025muon},
use cosine decay to \(0.01\) times peak learning rate, and report final target-step loss on C4-val, a fixed evaluation set drawn
from the C4 validation split. Hyperparameters are selected by target-step validation loss.

Seeds 42, 1234, and 5678 set the initialization and start the training
stream 0, 1024, and 2048 global batches in. At 360M, \(7.69\times\), the
offsets are 0, 128, and 256; at 1.3B, \(7.77\times\), every seed starts at
0, because the corpus ends just after the run. All 360M, 1.3B, and 3.1B
pretraining runs use GB300 GPUs. The 1.3B \(4\times\) and \(7.77\times\),
3.1B \(2\times\), and 360M \(16\times\) runs train on the 16-shard split of
C4 and are evaluated on the same C4-val tokens as the other runs.
Learning-rate brackets use seed 42. A pick on a bracket edge is extended
until it is interior, and the pick then gets three seeds. At 360M,
\(4\times\), Muon uses learning rate 0.002 and weight decay 0.3, which has
the lower three-seed mean of the two leading settings (2.4179 versus 2.4185
for 0.004), although seed 42 favors 0.004 by 0.2 milli-loss; the
\(4\times\) \(d\)-ladder reuses the directional learning rate 0.002.

\begin{table}[H]
    \centering
    \small
    \caption{Primary pretraining budgets.}
    \begin{tabular}{lrrr}
        \toprule
        Model & Budget & Steps & Train tokens \\
        \midrule
        360M & \(2\times\) & 13,732 & 14.399B \\
        360M & \(4\times\) & 27,464 & 28.798B \\
        360M & \(7.69\times\) & 52,800 & 55.365B \\
        1.3B & \(2\times\) & 48,584 & 50.944B \\
        1.3B & \(4\times\) & 97,168 & 101.888B \\
        1.3B & \(7.77\times\) & 188,739 & 197.907B \\
        3.1B & \(2\times\) & 119,605 & 125.415B \\
        \bottomrule
    \end{tabular}
\end{table}

Muon, MuonH, and the \(d=1\) control use separate brackets for the
matrix or directional update learning rate; Muon and dense AdamW additionally
tune weight decay in the core 360M experiments. In every Muon-family arm,
embedding/output-head parameters, normalization weights, and OrthMuon gain
parameters are trained with AdamW at a fixed learning rate of 0.0012, which we
do not tune in the main comparisons; Appendix~\ref{app:gain-lr} varies the
gain learning rate alone. The dense AdamW baseline instead tunes this as its main learning
rate. At both \(2\times\) and \(4\times\), the large-block SPD study in
Table~\ref{tab:large-d-full} separately selects the SPD-factor learning rate
as a multiple of the shared AdamW learning rate, holding the shared AdamW and
directional learning rates fixed. For
each \(d\in\{1,8,32\}\), a seed-42 directional-learning-rate bracket over
\(\{0.001,0.002,0.004\}\) selects the interior value 0.002, which is used for
the confirmation seeds. Factor-of-two spacing follows standard pretraining
optimizer benchmarks~\citep{wen2025fantastic}; because every block size selects
the same interior value, the bracket does not favor any \(d\).

\subsection{Model Architectures}

We evaluate three sizes of a Llama-style, decoder-only causal Transformer.
All models use pre-RMSNorm blocks, rotary positional embeddings (RoPE) with
base $\theta=10{,}000$, bias-free linear layers, and a SwiGLU feed-forward
network,
\[
\operatorname{MLP}(x)
=
W_{\mathrm{down}}
\left[
    \operatorname{SiLU}(W_{\mathrm{gate}}x)
    \odot
    W_{\mathrm{up}}x
\right].
\]
The feed-forward dimension is three times the model width. Input and output
token embeddings are tied, attention dropout is zero, and all pretraining
experiments use a vocabulary of 32,000 tokens and a maximum context length of
4,096 tokens. The model configurations are summarized in
Table~\ref{tab:model_architectures}.

\begin{table}[H]
\centering
\small
\caption{
Model architecture hyperparameters. The GQA ratio denotes the number of
query heads sharing each key--value head. Parameter counts refer to the base
Transformer with tied embeddings and exclude optimizer-specific auxiliary
parameters.
}
\label{tab:model_architectures}
\setlength{\tabcolsep}{4.5pt}
\begin{tabular}{lrrrrrrrl}
\toprule
Model &
Layers &
$d_{\mathrm{model}}$ &
$d_{\mathrm{ff}}$ &
Q heads &
KV heads &
Head dim. &
Parameters &
Attention \\
\midrule
360M   & 24 & 1024 & 3072 & 16 & 16 & 64  & 359.97M & MHA \\
1.3B   & 24 & 2048 & 6144 & 32 & 16 & 64  & 1.274B  & GQA ($2{:}1$) \\
3.1B & 32 & 2816 & 8448 & 22 & 11 & 128 & 3.135B  & GQA ($2{:}1$) \\
\bottomrule
\end{tabular}
\end{table}

The 360M model uses standard multi-head attention, with one key--value head
per query head. The 1.3B and 3.1B models use grouped-query attention:
each key--value head is shared by two query heads. Each Transformer layer
contains the four attention projections
$W_Q,W_K,W_V,W_O$ and the three SwiGLU projections
$W_{\mathrm{gate}},W_{\mathrm{up}},W_{\mathrm{down}}$.
Although these matrices may be represented in blocks for optimization, their
forward computation is identical to the corresponding dense linear
projections.

\section{OrthMuon Implementation Details}
\label{app:impl}

\paragraph{Blockwise update.}
Algorithm~\ref{alg:orthmuon-full} gives the update applied to each constrained
weight \(W\in\mathbb{R}^{m\times n}\) with row blocks of size \(d\leq n\).
Momentum, Newton--Schulz, and the scale are computed once for the whole matrix,
with \(G\) the gradient with respect to the stacked directional factors. The
tangent projection and the retraction then act on each block independently,
as a batched operation over the \(m/d\) blocks. For a tall matrix (\(m>n\)),
Newton--Schulz is applied to the transpose and the result is transposed back.
The five Newton--Schulz iterations are
\[
    X_{j+1}=\alpha X_j+\bigl(\beta X_jX_j^\top+\gamma(X_jX_j^\top)^2\bigr)X_j,
    \qquad (\alpha,\beta,\gamma)=(3.4445,-4.7750,2.0315),
\]
starting from \(X_0=M/\lVert M\rVert_F\).

\begin{algorithm}[h]
\caption{One OrthMuon step for a weight \(W\in\mathbb{R}^{m\times n}\) with
block size \(d\).}
\label{alg:orthmuon-full}
\small
\begin{tabular}{@{}r@{\hspace{0.6em}}p{0.62\textwidth}@{\hspace{0.5em}}>{\itshape\raggedright\arraybackslash}p{0.27\textwidth}@{}}
\toprule
\multicolumn{3}{@{}p{0.97\textwidth}@{}}{\textbf{Input:} blocks
\(Q_k\in\operatorname{St}(d,n)\) and gain factors \(D_k\) or \(L_k\),
\(k=1,\dots,m/d\); momentum \(B\); gradient \(G\); learning rate \(\eta\);
step \(t\); correction steps \(\mathcal{T}\), 50 evenly spaced steps.}\\
\midrule
1: & \(B\gets\mu B+G\), \(\;M\gets G+\mu B\) & Nesterov, \(\mu=0.95\)\\
2: & \(U\gets 0.2\sqrt{\max(m,n)}\,\operatorname{NS}_5\bigl(M/\lVert M\rVert_F\bigr)\) & whole matrix\\
3: & \textbf{for} \(k=1,\dots,m/d\) \textbf{in parallel do} & \\
4: & \quad\(A_k\gets Q_k-\eta\bigl(U_k-\operatorname{sym}(U_kQ_k^\top)Q_k\bigr)\) & tangent step\\
5: & \quad\(E_k\gets A_kA_k^\top-I_d\) \quad (fp32) & Gram error\\
6: & \quad\(Q_k\gets A_k+E_k\bigl(-\tfrac12I+E_k(\tfrac38I+E_k(-\tfrac5{16}I+\tfrac{35}{128}E_k))\bigr)A_k\) & Horner form\\
7: & \quad\textbf{if} \(t\in\mathcal{T}\) \textbf{then} \(Q_k\gets(Q_kQ_k^\top)^{-1/2}Q_k\) & exact polar factor\\
8: & update \(D_k\) or \(L_k\) with AdamW & Section~\ref{sec:recover}\\
\bottomrule
\end{tabular}
\end{algorithm}

\paragraph{Proof of Proposition~\ref{prop:retraction}.}
Tangency gives \(Q\Delta^\top+\Delta Q^\top=0\), so
\(AA^\top=I_m+\Delta\Delta^\top\) for \(A=Q+\Delta\). Hence
\(E=\Delta\Delta^\top\succeq0\) and \(\|E\|_2=\|\Delta\|_2^2\leq1\). Because
\(p_r(E)\) is a polynomial in \(E\),
\(Q^+Q^{+\top}=p_r(E)^2(I_m+E)=F_r(E)\) with \(F_r(x)=p_r(x)^2(1+x)\), and
\(\|Q^+Q^{+\top}-I_m\|_2=\max_\lambda|F_r(\lambda)-1|\) over the eigenvalues
\(\lambda\in[0,\|E\|_2]\) of \(E\). Write
\(p_r(x)=\sum_{k=0}^r\beta_kx^k\) with \(\beta_k=\binom{-1/2}{k}\). The
function \(u=(1+x)^{-1/2}\) satisfies \(2(1+x)u'+u=0\), so its coefficients
obey \(2(k+1)\beta_{k+1}+(2k+1)\beta_k=0\). Truncation leaves only the top
term:
\[
    2(1+x)p_r'(x)+p_r(x)=(2r+1)\beta_rx^r=-2(r+1)\beta_{r+1}x^r.
\]
Hence \(F_r'=p_r\bigl(2(1+x)p_r'+p_r\bigr)=-2(r+1)\beta_{r+1}x^rp_r(x)\),
and since \(F_r(0)=1\),
\[
    F_r(x)-1=-2(r+1)\beta_{r+1}\int_0^x t^r p_r(t)\,\mathrm{d}t.
\]
On \([0,1]\) the series of \(u\) alternates with terms of decreasing
magnitude, so its partial sums satisfy \(\tfrac12\leq1-\tfrac t2\leq
p_r(t)\leq1\) for \(r\geq1\). Therefore
\(|F_r(x)-1|\leq2|\beta_{r+1}|x^{r+1}\), with sign \((-1)^r\) and equality
to leading order because \(p_r(0)=1\). Since
\(|\beta_{r+1}|=\binom{2r+2}{r+1}/4^{r+1}\), this constant is \(c_r\), and
substituting \(\lambda\leq\|\Delta\|_2^2\) gives the bound. For \(r=4\), the
leading term is \(Q^+Q^{+\top}-I_m=\tfrac{63}{128}E^5+O(E^6)\).

For the cost, Horner evaluation writes
\(p_r(E)A=A+E\bigl(\beta_1I+E(\beta_2I+\cdots+E\beta_r)\bigr)A\). It needs
the Gram product \(AA^\top\) (\(2m^2n\) FLOPs), \(r-1\) products of
\(m\times m\) matrices (\(2m^3\) each), and one product with \(A\)
(\(2m^2n\)), and it never multiplies by the identity. \(\square\)

\paragraph{Step size and the choice \(r=4\).}
Newton--Schulz starts from \(X_0=M/\|M\|_F\), whose singular values lie in
\([0,1]\), and maps each of them through the same scalar quintic; on
\([0,1]\), five iterations stay below 1.203, so \(\|Z\|_2\leq1.203\) in exact
arithmetic. The projection splits \(Z\) into
\(Z(I-Q^\top Q)\) and \(\operatorname{skew}(ZQ^\top)Q\), whose rows lie in
orthogonal subspaces, so \(\|\Pi_Q(Z)\|_2\leq\sqrt2\|Z\|_2\). Hence
\(\|\Delta\|_2\leq1.71\,\eta s\). By Proposition~\ref{prop:retraction}, the
retraction error stays below the fp32 unit roundoff \(2^{-24}\) whenever
\[
    \eta s\leq 0.040\ (r=2),\qquad 0.079\ (r=3),\qquad 0.120\ (r=4).
\]
For a \(1024\times1024\) matrix at learning rate 0.002,
\(\eta s=0.0128\). Raising \(r\) from two to four costs two extra
\(d\times d\) products per block, \(4md^2\) FLOPs per matrix, but triples
the step size covered by this guarantee. The implementation originally chose
\(r\in\{2,3,4\}\), or an exact polar step, from \(\|E\|_F\). That choice
reads a scalar back to the host at every step, which synchronizes the GPU
with the CPU and breaks graph compilation. Fixing \(r=4\) removes the
synchronization, and the projection and retraction compile into one graph.

\paragraph{Implementation of the retraction.}
The polar factor of \(p_r(E)A\) equals that of \(A\), because \(p_r(E)\) is
symmetric positive definite and commutes with \(AA^\top\). An exact
correction applied after the polynomial therefore yields exactly the polar
retraction of the tangent step. When the parameters are stored in bfloat16,
the Gram error and the polynomial are computed in fp32 and the result is cast
back. Exact corrections compute the inverse square root from an
eigendecomposition of \(QQ^\top\), with an SVD fallback. They occur at 50
evenly spaced steps regardless of the training length.

\paragraph{Arithmetic cost.}
Counting matrix-multiply FLOPs, one Newton--Schulz iteration on an
\(m\times n\) matrix with \(m\leq n\) costs \(4m^2n+2m^3\), so five
iterations cost \(20m^2n+10m^3\). For one \(d\times n\) block, the tangent
projection costs \(4d^2n\) (the products \(UQ^\top\) and
\(\operatorname{sym}(\cdot)Q\)), and the order-4 retraction costs
\(4d^2n+6d^3\) by Proposition~\ref{prop:retraction}. Over all \(m/d\) blocks,
the added cost is \(8mdn+6md^2\), at most \(0.6\,d/m\) of the Newton--Schulz
cost when \(m\leq n\). For \(m>n\), the same count gives at most
\(0.7\,d/n\). For a \(1024\times1024\) projection with \(d=32\), the added
cost is 0.9\% of Newton--Schulz.
Measured throughput does not grow with \(d\)
(Table~\ref{tab:throughput}): with per-row gains, OrthMuon is 1.3--1.8\%
slower than Muon at 360M and at most 1.6\% slower at 1.3B.

\begin{table}[H]
    \centering
    \small
    \caption{Training throughput (thousand tokens/s) on 8 GB300 GPUs at the
    \(2\times\) budget: median over all runs of each method (per-run median
    after step 500). Change is relative to Muon at the same scale.}
    \label{tab:throughput}
    \setlength{\tabcolsep}{3pt}
    \begin{tabular}{lrrrr}
        \toprule
        & \multicolumn{2}{c}{360M} & \multicolumn{2}{c}{1.3B} \\
        \cmidrule(lr){2-3}\cmidrule(lr){4-5}
        Method & k tokens/s & Change & k tokens/s & Change \\
        \midrule
        Muon & 2134 & -- & 870.3 & -- \\
        OrthMuon \(d=1\) & 2095 & \(-1.8\%\) & 856.2 & \(-1.6\%\) \\
        OrthMuon \(d=8\) & 2095 & \(-1.8\%\) & -- & -- \\
        OrthMuon \(d=32\) & 2107 & \(-1.3\%\) & 871.0 & \(+0.1\%\) \\
        \bottomrule
    \end{tabular}
\end{table}

\paragraph{Parameters and initialization.}
The constrained matrices are the \(q,k,v,o\) projections and the MLP gate, up,
and down projections. Embeddings, embedding-tied output parameters, and
normalization weights remain in AdamW. Matrix rows are output channels, except that the MLP down
projection is stored as an MLP-width-by-model-width matrix (\(3072\times1024\)
at 360M) whose rows are its input channels; constraining its output-side rows
instead gives the same loss. Rows are grouped contiguously; a row count not divisible by \(d\) raises an error
rather than creating a remainder block. Each directional factor is
initialized by orthonormalizing a standard random initialization with a
reduced QR decomposition. Matrix-gain runs initialize the effective left gain
to \(S_k=I_d\). Row gains and \(L_k\) are trained with AdamW at the shared
AdamW-side learning rate 0.0012 and zero weight decay; deviations are stated
with the corresponding ablation. Constrained directional factors receive no
weight decay.

\paragraph{Manifold-Muon solvers.}
At full width (\(d\) equal to the number of matrix rows), manifold Muon uses
the exact square-case update
\(\Phi=Q\,\operatorname{msign}(\operatorname{skew}(Q^\top G))\), where
\(\operatorname{skew}(A)=\tfrac12(A-A^\top)\) and
\(\operatorname{msign}(A)=UV^\top\) for an SVD \(A=U\Sigma V^\top\).
For \(d\leq256\), it instead uses a thin-QR reduction to a
\(2d\times2d\) matrix. The gain parameterization and retraction are unchanged.

\paragraph{Direction--magnitude coordinates.}
Even when two parameterizations represent the same effective weights, they
need not induce the same update. For one row, write \(w=aq\) with
\(\lVert q\rVert_2=1\). An infinitesimal effective update decomposes as
\begin{equation}
    \mathrm{d}w=q\,\mathrm{d}a+a\,\mathrm{d}q,
    \qquad q^\top\mathrm{d}q=0.
\end{equation}
The gain controls both the effective row norm and the conversion between a
step in \(q\) and an angular step in \(w\).

\paragraph{Polar expressivity.}
For a full-row-rank block \(W_k\), the matrix
\(P_k=(W_kW_k^\top)^{1/2}\) is positive definite, and
\begin{equation}
    W_k=P_kQ_k,\qquad
    P_k=(W_kW_k^\top)^{1/2},\qquad
    Q_k=P_k^{-1}W_k,
\end{equation}
where \(Q_kQ_k^\top=I_d\). Thus a positive-definite matrix gain can
represent any full-row-rank effective block even though the latent directional
factor remains orthogonal.

\section{Other Optimizers and Controls}
\label{app:baselines}

\paragraph{Manifold Muon.}
Manifold Muon solves spectral-norm steepest descent directly inside the
Stiefel tangent space~\citep{bernstein2025manifolds,su2025stiefel,solonko2026stiefel},
with the same magnitude parameterization and retraction as OrthMuon;
Appendix~\ref{app:impl} gives the full-width and thin-QR solvers.

\paragraph{Forward normalization.}
The same effective function class need not induce the same update. For
\(w=aq\), the gain couples a step in \(q\) to the effective angular step
(Appendix~\ref{app:impl}). Forward normalization stores momentum on an
unconstrained latent weight, applies Muon to it, and differentiates through
row normalization; it preserves the \(d=1\) function class but changes the
optimization trajectory. The default OrthMuon stores momentum in ambient
coordinates; tangent and transported-momentum variants serve as further
controls (Appendix~\ref{app:optimizer-controls}).

\paragraph{No-gain control.}
Setting \(D_k=I_d\) removes learned magnitudes. At \(d=1\), this fixes every
row norm without constraining angles between rows. Dense Muon, no-gain
\(d=1\), and learned-gain \(d=1\) form our norm-and-magnitude ablation.

\paragraph{MuonH/Hyperball.}
MuonH constrains the Frobenius norm of each full matrix to its initial value,
explicitly normalizes its update, and uses no learned gain or weight
decay~\citep{wen2026hyperball}. It is a coarser norm-control baseline with no
pairwise orthogonality.

\paragraph{Other left factors.}
Besides the diagonal and SPD factors of Section~\ref{sec:recover}, we train
eight other left factors \(W_k=F_kQ_k\), all initialized at \(F_k=I_d\):
\(F_k=X_kX_k^\top\) with a full square \(X_k\) (XXT);
\(\operatorname{diag}(a_k)+X_kX_k^\top\), whose \(X_k\) is updated with Muon;
the symmetric square of a symmetric matrix, also updated with Muon;
\(L_kL_k^\top+\operatorname{diag}(\lambda_k)\); the matrix exponential
\(\exp(A_k)\) of a symmetric \(A_k\); an indefinite symmetric matrix
\(\operatorname{diag}(a_k)+\operatorname{Off}(M_k)\); a lower-triangular
\(T_k\) with diagonal \(\operatorname{diag}(a_k)\), which gives the LQ form
\(W_k=T_kQ_k\); and a singular-value form \(U_k\operatorname{diag}(s_k)\),
whose orthogonal \(U_k\) is updated with the same Stiefel update as \(Q_k\).
Except where noted, the factors are trained with AdamW. Every one of them can
represent any full-row-rank block. Results are in Appendix~\ref{app:gains}.

\begin{table}[H]
    \centering
    \small
    \caption{The fourteen orthogonality-training variants, the three
    alternative updates for the sphere control (rows 12--14), and where their
    results appear.}
    \label{tab:variants}
    \begin{tabular}{rll}
        \toprule
        & Variant & Results \\
        \midrule
        \multicolumn{3}{l}{\emph{OrthMuon with block-Stiefel directions; left factor \(F_k\) in \(W_k=F_kQ_k\)}} \\
        1 & Per-row gain \(\operatorname{diag}(a_k)\) & Table~\ref{tab:main}, Figure~\ref{fig:blocksize} \\
        2 & One scalar gain per block & Table~\ref{tab:dgreater1-attempts} \\
        3 & SPD \(L_kL_k^\top\) & Tables~\ref{tab:dgreater1-attempts}, \ref{tab:gain-param} \\
        4 & \(X_kX_k^\top\) & Tables~\ref{tab:dgreater1-attempts}, \ref{tab:gain-param} \\
        5 & \(\operatorname{diag}(a_k)+X_kX_k^\top\), \(X_k\) updated with Muon & Table~\ref{tab:dgreater1-attempts} \\
        6 & Symmetric square, updated with Muon & Table~\ref{tab:matrix-gain} \\
        7 & \(L_kL_k^\top+\operatorname{diag}(\lambda_k)\) & Table~\ref{tab:gain-param} \\
        8 & Matrix exponential \(\exp(A_k)\) & Table~\ref{tab:gain-param} \\
        9 & Indefinite symmetric \(\operatorname{diag}(a_k)+\operatorname{Off}(M_k)\) & Table~\ref{tab:gain-param} \\
        10 & Lower triangular \(T_k\) & Table~\ref{tab:gain-param} \\
        11 & Singular-value \(U_k\operatorname{diag}(s_k)\) & Table~\ref{tab:gain-param} \\
        \midrule
        \multicolumn{3}{l}{\emph{Alternative updates for \(d=1\)}} \\
        12 & Direct row normalization & Table~\ref{tab:d1-controls} \\
        13 & Tangent gradient before momentum & Table~\ref{tab:d1-controls} \\
        14 & Tangent gradient with transported momentum & Table~\ref{tab:d1-controls} \\
        \midrule
        \multicolumn{3}{l}{\emph{Other optimizers for the same constraint}} \\
        15 & Manifold Muon, per-row gain & Table~\ref{tab:manifold-muon-360} \\
        16 & Manifold Muon, SPD gain & Table~\ref{tab:manifold-muon-360} \\
        17 & Projected diagonal-Gram (DGram) & Table~\ref{tab:dgreater1-attempts}; App.~\ref{app:dgram} \\
        \bottomrule
    \end{tabular}
\end{table}

\section{Adaptation Setup}

For a base weight \(W_0\in\mathbb{R}^{p\times q}\), StelLA writes the
rank-\(r\) update as~\citep{li2025stella}
\begin{equation}
    \Delta W=\frac{\alpha}{r}USV^\top,
    \qquad
    U\in\mathbb{R}^{p\times r},\
    S\in\mathbb{R}^{r\times r},\
    V\in\mathbb{R}^{q\times r}.
    \label{eq:stella}
\end{equation}
The rank directions are the columns of \(U\) and \(V\). Euclidean factors are
unconstrained. Sphere optimizes each rank direction independently on a unit
sphere while allowing arbitrary pairwise angles. Stiefel additionally requires
\(U^\top U=V^\top V=I_r\). All three use the same learned \(S\), making
Stiefel versus Sphere the direct comparison for adding pairwise angle
constraints at fixed factor norms. The canonical Stiefel update
tangent-projects both the ambient gradient and the post-Adam displacement
before polar retraction; the direct-polar control omits only the post-Adam
tangent projection.

The adaptation study uses Llama-3-8B with rank 32, \(\alpha=64\), and
adapters on the query, key, value, up-projection, and down-projection
matrices~\citep{grattafiori2024llama}. We train for three epochs with global
batch 16, learning rate \(5\times10^{-4}\), linear decay, and 500 warmup
steps. All main-table arms use StelLA's default learning rate
\(5\times10^{-4}\) and \(S_0=I\); each run loads the checkpoint with the lowest
token-level loss on a fixed 120-example validation split. The learning-rate
and decay grids in Appendix~\ref{app:stella} are sensitivity checks, not
selection rules for the main table. Scores are equal-weight macro accuracy
over BoolQ, PIQA, SocialIQA, HellaSwag, WinoGrande, ARC-Easy, ARC-Challenge,
and OpenBookQA.

\section{Seed Counts and Uncertainty}
\label{app:seeds}

Main-text tables and text report means over seeds.
Independent-seed replication is
uncommon at this scale in pretraining comparisons; for example, the 1B--8B
comparisons of \citet{lion2026muown} and \citet{wen2026hyperball} use a single
run per method. Every 360M, 1.3B, and 3.1B pretraining cell uses three
seeds, each with its own initialization and data start
(Appendix~\ref{app:protocols}). The exceptions are the 360M \(16\times\) run
(one seed; Appendix~\ref{app:overtrain}) and 1.3B \(d=1024\) SPD (two
converged seeds; seed 42 diverged; Table~\ref{tab:large-d-1p3b}).
Learning-rate brackets and sensitivity sweeps use seed 42, except the
projected-DGram learning-rate and weight-decay grid, which uses two seeds
(Appendix~\ref{app:dgram}).
The diffusion-transformer comparison uses five seeds (42--46), and its
query/key SPD run one (Appendix~\ref{app:dit}); the fine-tuning study uses the
three pretraining seeds of each base (Appendix~\ref{app:sft}).

\paragraph{Seed counts in the main text.}
\begin{itemize}
    \item Table~\ref{tab:precision}: one seed per cell; the POET entries use
    the best learning rate of a per-precision sweep, and the Pion entries use
    seed 42.
    \item Figure~\ref{fig:stella}: four seeds (42--45).
\end{itemize}

\paragraph{Paired intervals and dispersion.}
In the matched ladder, \(d=8\) changes loss with mixed signs across seeds at
\(2\times\) and \(4\times\), whereas \(d=32\) is worse than \(d=1\) on every
seed at all three budgets (Table~\ref{tab:paired-d-effects}). The lower ends
of the \(d=8-d=1\) intervals rule out an improvement larger than 4.9
milli-loss at \(2\times\) and 0.3 at \(7.69\times\). At 1.3B, \(d=32\) SPD
minus \(d=1\) is \(-0.98\), with 95\% interval \([-2.25,0.29]\). In
StelLA, the Sphere-over-Euclidean gain has 95\% paired
interval \([1.73,2.50]\) and the Stiefel-minus-Sphere change has
\([-0.19,0.01]\) (Appendix~\ref{app:stella}). In PSOFT, the row-sphere
advantage over PSOFT at the selected learning rates has paired interval
\([-0.28,1.33]\), and the advantage over the dense arm has \([-0.29,0.53]\)
(Appendix~\ref{app:psoft}).

\paragraph{Bounds on improvement.}
Because our claim is the absence of a gain, the relevant quantity is the
largest improvement over \(d=1\) that each 95\% interval still allows.
Per-row \(d=8\) cannot improve on \(d=1\) by more than 4.9, 2.3, and 0.3
milli-loss at \(2\times\), \(4\times\), and \(7.69\times\); per-row \(d=32\)
is worse at every budget, and no interval allows an improvement. SPD \(d=32\)
cannot improve by more than 4.6 milli-loss at 360M, 2.3 at 1.3B (1.1 against
the gain-tuned \(d=1\)), and 2.5 at 3.1B. Over the main ladder we make six
\(d>1\) contrasts (two block sizes at three budgets) and do not adjust for
multiplicity; after a Bonferroni correction for these six contrasts, \(d=32\)
remains worse at \(4\times\) and \(7.69\times\).

\section{Per-Seed Matched Orthogonality Ladders}
\label{app:d-ladder}

\begin{table}[H]
    \centering
    \small
    \caption{Paired change in C4 validation loss from adding row orthogonality,
    relative to the \(d=1\) control, in milli-loss (positive means
    worse). Each difference compares the same seed; intervals are paired 95\%
    \(t\)-intervals, and the final column counts higher-loss \(d>1\) runs.}
    \label{tab:paired-d-effects}
    \begin{tabular}{llrrr}
        \toprule
        Budget & Contrast & Mean \(\Delta\) & 95\% interval & Worse seeds \\
        \midrule
        \(2\times\) & \(d=8-d=1\) & +0.81 & \([-4.89,6.50]\) & 2/3 \\
        \(2\times\) & \(d=32-d=1\) & +5.19 & \([2.54,7.83]\) & 3/3 \\
        \(4\times\) & \(d=8-d=1\) & +0.87 & \([-2.25,3.99]\) & 2/3 \\
        \(4\times\) & \(d=32-d=1\) & +5.09 & \([3.53,6.65]\) & 3/3 \\
        \(7.69\times\) & \(d=8-d=1\) & +1.10 & \([-0.33,2.54]\) & 3/3 \\
        \(7.69\times\) & \(d=32-d=1\) & +3.71 & \([2.79,4.63]\) & 3/3 \\
        \bottomrule
    \end{tabular}
\end{table}

\begin{table}[H]
    \centering
    \small
    \caption{Per-seed final C4 loss for the matched per-row-gain ladder.}
    \label{tab:d-ladder-seeds}
    \begin{tabular}{crrrrr}
        \toprule
        Budget & \(d\) & Seed 42 & Seed 1234 & Seed 5678 & Mean \(\pm\) SD \\
        \midrule
        \(2\times\) & 1  & 2.4616 & 2.4631 & 2.4617 & 2.4621 \(\pm\) 0.0009 \\
        \(2\times\) & 8  & 2.4637 & 2.4613 & 2.4639 & 2.4629 \(\pm\) 0.0014 \\
        \(2\times\) & 32 & 2.4660 & 2.4679 & 2.4681 & 2.4673 \(\pm\) 0.0012 \\
        \midrule
        \(4\times\) & 1  & 2.3999 & 2.4001 & 2.4006 & 2.4002 \(\pm\) 0.0004 \\
        \(4\times\) & 8  & 2.4018 & 2.3995 & 2.4018 & 2.4011 \(\pm\) 0.0013 \\
        \(4\times\) & 32 & 2.4043 & 2.4057 & 2.4058 & 2.4053 \(\pm\) 0.0008 \\
        \midrule
        \(7.69\times\) & 1  & 2.3557 & 2.3563 & 2.3572 & 2.3564 \(\pm\) 0.0007 \\
        \(7.69\times\) & 8  & 2.3575 & 2.3571 & 2.3579 & 2.3575 \(\pm\) 0.0004 \\
        \(7.69\times\) & 32 & 2.3598 & 2.3598 & 2.3607 & 2.3601 \(\pm\) 0.0005 \\
        \bottomrule
    \end{tabular}
\end{table}

All rows in Table~\ref{tab:d-ladder-seeds} use one learned gain per weight
row. Main-text differences are computed from the unrounded values. The paired
\(d=32-d=1\) effects are as in Table~\ref{tab:paired-d-effects}: \(+5.19\),
\(+5.09\), and \(+3.71\) milli-loss at \(2\times\), \(4\times\), and
\(7.69\times\), each worse on every seed. The \(d=8-d=1\) signs are mixed at
\(2\times\) and \(4\times\). At \(7.69\times\), \(d=8\) is higher on every
seed, but its interval includes zero.

\section{Training Past \texorpdfstring{\(7.69\times\)}{7.69x}}
\label{app:overtrain}

\begin{table}[H]
    \centering
    \small
    \caption{Per-row \(d=32\) against \(d=1\) at 360M from \(2\times\) to
    \(16\times\) (gap and drop in milli-loss): the loss gap shrinks, but the
    extra training that \(d=32\) needs stays near 5--6\%.}
    \setlength{\tabcolsep}{4pt}
    \begin{tabular}{lccccc}
        \toprule
        & & & & \(d=1\) drop per & Extra steps \\
        Budget & \(d=1\) & Per-row \(d=32\) & \(d=32-d=1\) & step doubling & \(d=32\) needs \\
        \midrule
        \(2\times\) & 2.4621 & 2.4673 & \(+5.19\) & 62.0 (\(2\times\to4\times\)) & 6.0\% \\
        \(4\times\) & 2.4002 & 2.4053 & \(+5.09\) & 46.4 (\(4\times\to7.69\times\)) & -- \\
        \(7.69\times\) & 2.3564 & 2.3601 & \(+3.71\) & -- & 5.7\% \\
        \(16\times\) & 2.3200 & 2.3224 & \(+2.44\) & 34.5 (\(7.69\times\to16\times\)) & 5.0\% \\
        \bottomrule
    \end{tabular}
\end{table}

The \(2\times\) to \(7.69\times\) rows are three-seed means; the \(16\times\)
row is seed 42. The drop per step doubling divides the \(d=1\) loss decrease
between two budgets by the base-2 logarithm of their step ratio. Extra steps
is \(2^{g/r}-1\), the step increase at which \(d=32\) would match \(d=1\) if
its loss fell at the listed \(d=1\) rate \(r\), where \(g\) is the loss gap.
The step cost stays near 5--6\%; the loss gap shrinks because the loss
curve flattens. The \(16\times\) run (109,856
steps, 115.2B tokens) is one seed and trains on the 16-shard split of C4; its
validation tokens are the same as at the other budgets.

\section{Sphere Optimization and Learned Magnitude}

\begin{table}[H]
    \centering
    \small
    \caption{Final C4 validation loss at 360M (three-seed means).}
    \begin{tabular}{lccc}
        \toprule
        Method & \(2\times\) & \(4\times\) & \(7.69\times\) \\
        \midrule
        Tuned Muon & 2.475 & 2.418 & 2.372 \\
        MuonH & 2.469 & 2.407 & 2.362 \\
        \(d=1\), no gain & 2.468 & 2.408 & 2.365 \\
        \(d=1\), row gain & \textbf{2.462} & \textbf{2.400} & \textbf{2.356} \\
        \bottomrule
    \end{tabular}
\end{table}

\begin{table}[H]
    \centering
    \small
    \caption{Loss improvements over tuned Muon at 360M in milli-loss, computed
    from unrounded means, so displayed components need not add exactly to the
    displayed total; positive values indicate lower loss.}
    \begin{tabular}{lccc}
        \toprule
        Component & \(2\times\) & \(4\times\) & \(7.69\times\) \\
        \midrule
        Unit-row direction constraint, no gain & 7.2 & 10.1 & 7.1 \\
        Learned row magnitude & 5.9 & 7.5 & 8.6 \\
        Total \(d=1\) improvement & 13.1 & 17.7 & 15.7 \\
        \bottomrule
    \end{tabular}
\end{table}

MuonH is within 0.6 milli-loss of no-gain \(d=1\) at \(2\times\) and
\(4\times\) (\(+0.5\) and \(-0.6\)), despite constraining one norm per matrix
versus one norm per row, and 2.7 milli-loss better at \(7.69\times\).

\section{Scaling of the \(d=1\) Control}
\label{app:scaling}

\begin{table}[H]
    \centering
    \small
    \caption{Scaling of the \(d=1\) control on C4 (three-seed means). The
    last two columns give the paired gap to \(d=1\) in milli-loss, as mean
    \(\pm\) 95\% half-width.}
    \setlength{\tabcolsep}{4pt}
    \begin{tabular}{llcccrr}
        \toprule
        Model & Budget & Muon & MuonH & \(d=1\) & Muon \(-\,d=1\) & MuonH \(-\,d=1\) \\
        \midrule
        1.3B & \(2\times\) & 2.1970 & 2.1911 & \textbf{2.1842} & \(+12.81\pm1.50\) & \(+6.86\pm1.07\) \\
        1.3B & \(4\times\) & 2.1524 & 2.1464 & \textbf{2.1393} & \(+13.08\pm2.45\) & \(+7.08\pm0.55\) \\
        1.3B & \(7.77\times\) & 2.1198 & 2.1141 & \textbf{2.1079} & \(+11.97\pm1.30\) & \(+6.26\pm1.37\) \\
        3.1B & \(2\times\) & 2.0362 & 2.0307 & \textbf{2.0242} & \(+12.03\pm1.83\) & \(+6.57\pm0.20\) \\
        \bottomrule
    \end{tabular}
\end{table}

\begin{table}[H]
    \centering
    \small
    \caption{Larger-scale \(d=32\) comparisons on C4 (three-seed means).
    \(\Delta\) is the paired difference from the \(d=1\) result at the same
    model size, in milli-loss, as mean \(\pm\) 95\% half-width.}
    \begin{tabular}{lrcr}
        \toprule
        Model & \(d\) / gain & Loss & \(\Delta\times10^3\) \\
        \midrule
        1.3B & \(d=1\), per row & 2.1842 & -- \\
        1.3B & \(d=32\), per row & 2.1861 & \(+1.92\pm1.52\) \\
        1.3B & \(d=32\), SPD & 2.1832 & \(-0.98\pm1.27\) \\
        \midrule
        3.1B & \(d=1\), per row & 2.0242 & -- \\
        3.1B & \(d=32\), per row & 2.0249 & \(+0.78\pm0.92\) \\
        3.1B & \(d=32\), SPD & 2.0239 & \(-0.28\pm2.25\) \\
        \bottomrule
    \end{tabular}
\end{table}

\section{FPO-Mix Training-Mixture Check}
\label{app:fpo}

FPO-mix denotes the pretraining mixture of \citet{wen2025fantastic}. It contains 54.401B tokens, with
realized proportions 92.57\% DCLM, 6.09\% StarCoder, and 1.34\% ProofPile.
Training-mixture comparisons use C4-val.

\begin{table}[H]
    \centering
    \small
    \caption{Seed-42 FPO-mix learning-rate sweep at 360M, \(2\times\).}
    \begin{tabular}{lrr}
        \toprule
        Method & LR & Final C4 loss \\
        \midrule
        MuonH & 0.00625 & 2.7014 \\
        MuonH & 0.0125 & \textbf{2.6759} \\
        MuonH & 0.025 & 2.7140 \\
        OrthMuon \(d=1\) & 0.001 & 2.7075 \\
        OrthMuon \(d=1\) & 0.002 & \textbf{2.6695} \\
        OrthMuon \(d=1\) & 0.004 & 2.6818 \\
        \bottomrule
    \end{tabular}
\end{table}

\begin{table}[H]
    \centering
    \small
    \caption{Three-seed FPO-mix results at 360M, \(2\times\), evaluated on C4-val.}
    \begin{tabular}{lcc}
        \toprule
        Method & \(n\) & Mean loss \(\pm\) SD \\
        \midrule
        OrthMuon \(d=1\) & 3 & \textbf{2.6709} \(\pm\) 0.0014 \\
        OrthMuon \(d=32\), per-row gain & 3 & 2.6753 \(\pm\) 0.0004 \\
        MuonH & 3 & 2.6771 \(\pm\) 0.0011 \\
        Muon & 3 & 2.6824 \(\pm\) 0.0013 \\
        \bottomrule
    \end{tabular}
\end{table}

The Muon row in the preceding three-seed table uses learning rate 0.004 and
weight decay 0.3, selected by a seed-42 bracket over learning rates
\(\{0.002,0.004,0.008\}\).

\begin{table}[H]
    \centering
    \small
    \caption{360M FPO-mix \(d=1\) and \(d=32\) SPD at the shared AdamW-side
    learning rate 0.0012, evaluated on C4-val. Entries at directional learning
    rate 0.002 are three-seed means; SPD entries at 0.001 and 0.004 are the
    seed-42 bracket.}
    \begin{tabular}{clrr}
        \toprule
        Budget & Method & Directional LR & Final loss \\
        \midrule
        \(2\times\) & \(d=1\) & 0.002 & 2.6709 \\
        \(2\times\) & \(d=32\) SPD & 0.001 & 2.7057 \\
        \(2\times\) & \(d=32\) SPD & 0.002 & 2.6705 \\
        \(2\times\) & \(d=32\) SPD & 0.004 & 2.6806 \\
        \midrule
        \(4\times\) & \(d=1\) & 0.002 & 2.6078 \\
        \(4\times\) & \(d=32\) SPD & 0.001 & 2.6264 \\
        \(4\times\) & \(d=32\) SPD & 0.002 & 2.6071 \\
        \(4\times\) & \(d=32\) SPD & 0.004 & 2.6246 \\
        \bottomrule
    \end{tabular}
\end{table}

At the common directional learning rate 0.002, SPD minus \(d=1\) is
\(-0.44\) at \(2\times\) and \(-0.70\) at \(4\times\) (95\% half-widths
1.58 and 1.93). These are training-mixture checks; the reported metric
remains C4 validation loss.

\begin{table}[H]
    \centering
    \small
    \caption{FPO-mix 1.3B, \(2\times\), evaluated on C4-val.}
    \begin{tabular}{lccc}
        \toprule
        Method & Selected LR / WD & \(n\) & Mean loss \(\pm\) SD \\
        \midrule
        OrthMuon \(d=1\) & 0.001 & 3 & \textbf{2.4004} \(\pm\) 0.0005 \\
        MuonH & 0.008 & 3 & 2.4071 \(\pm\) 0.0001 \\
        Muon & 0.002 / 0.3 & 3 & 2.4127 \(\pm\) 0.0003 \\
        \bottomrule
    \end{tabular}
\end{table}

\section{Alternate-Corpus Evaluation}
\label{app:ood}

We evaluate the final 360M C4-trained checkpoints at \(2\times\) and
\(7.69\times\) on six alternate-corpus sets without further training, at sequence
length 4096, using 12.6M tokens per FPO set and 14.7M tokens per OLM set.
Every checkpoint was also replayed on C4; the maximum deviation from its
recorded loss was at most 0.0018 at both budgets. C4
removed every page containing a curly bracket~\citep{raffel2020exploring},
which biases it away from code- and LaTeX-heavy pages. We therefore report the
predominantly natural-language sets (FPO mixture, DCLM, OLM-Wiki) separately
from StarCoder, ProofPile, and OLM-arXiv.

Every arm has three seeds (Tables~\ref{tab:ood-2x} and \ref{tab:ood-7p69-matched}), and losses
off C4 vary much more across seeds than on C4 (Table~\ref{tab:ood-noise}): the three \(d=1\) seeds
span 1.5 milli-loss on C4 but 7--9 on the FPO mixture and DCLM and 59--106 on StarCoder, ProofPile,
and OLM-arXiv. We therefore compare arms by seed-paired differences from \(d=1\)
(Table~\ref{tab:ood-paired}). On the FPO mixture and DCLM, the C4 ordering carries over at about its
C4 magnitude: per-row \(d=8\) and \(d=32\) are within noise of \(d=1\), Muon and MuonH trail it by 7
to 14 milli-loss, and AdamW by about 90; across arms, these differences correlate with the C4
differences at \(r=0.70\) to 0.85 (AdamW excluded). OLM-Wiki is intermediate. On StarCoder,
ProofPile, and OLM-arXiv, most paired intervals are wider than \(\pm80\) milli-loss, and three seeds
do not rank the arms; at \(2\times\), Muon is nominally better than \(d=1\) on ProofPile
(\(-87\pm83\)), a possible trade-off that these data do not establish.

\begin{table}[H]
    \centering
    \scriptsize
    \caption{Mean loss of 360M, \(2\times\), C4-trained checkpoints on alternate-corpus
    sets (three seeds).}
    \label{tab:ood-2x}
    \setlength{\tabcolsep}{3pt}
    \begin{tabular}{lccccccc}
        \toprule
        & & \multicolumn{3}{c}{Natural-language-heavy} & \multicolumn{3}{c}{Code/math/LaTeX-rich} \\
        \cmidrule(lr){3-5}\cmidrule(lr){6-8}
        Method & C4 & FPO mix & DCLM & OLM-Wiki & StarCoder & ProofPile & OLM-arXiv \\
        \midrule
        AdamW & 2.5511 & 3.0992 & 3.0984 & 3.2295 & 3.0594 & 3.9805 & 2.8222 \\
        Muon & 2.4752 & 3.0203 & 3.0208 & 3.1395 & 2.9839 & 3.7714 & 2.7372 \\
        MuonH & 2.4686 & 3.0197 & 3.0158 & 3.1452 & 3.0407 & 3.8348 & 2.7597 \\
        \(d=1\), no gain & 2.4681 & 3.0162 & 3.0161 & 3.1461 & 2.9789 & 3.8657 & 2.7456 \\
        \(d=1\) & 2.4621 & 3.0113 & 3.0070 & 3.1254 & 3.0315 & 3.8587 & 2.7449 \\
        \(d=8\), per row & 2.4629 & 3.0092 & 3.0061 & 3.1345 & 3.0044 & 3.8527 & 2.7466 \\
        \(d=32\), per row & 2.4673 & 3.0160 & 3.0149 & 3.1301 & 2.9961 & 3.8313 & 2.7665 \\
        \bottomrule
    \end{tabular}
\end{table}

\begin{table}[H]
    \centering
    \scriptsize
    \caption{Mean loss of 360M, \(7.69\times\), C4-trained checkpoints on
    alternate-corpus sets (three seeds).}
    \label{tab:ood-7p69-matched}
    \setlength{\tabcolsep}{3pt}
    \begin{tabular}{lccccccc}
        \toprule
        & & \multicolumn{3}{c}{Natural-language-heavy} & \multicolumn{3}{c}{Code/math/LaTeX-rich} \\
        \cmidrule(lr){3-5}\cmidrule(lr){6-8}
        Method & C4 & FPO mix & DCLM & OLM-Wiki & StarCoder & ProofPile & OLM-arXiv \\
        \midrule
        Muon & 2.3721 & 2.9399 & 2.9390 & 3.0258 & 2.9310 & 3.7282 & 2.6618 \\
        MuonH & 2.3623 & 2.9340 & 2.9353 & 3.0572 & 2.8948 & 3.7337 & 2.6993 \\
        \(d=1\), no gain & 2.3650 & 2.9286 & 2.9345 & 3.0460 & 2.8328 & 3.7094 & 2.6817 \\
        \(d=1\) & 2.3564 & 2.9266 & 2.9284 & 3.0390 & 2.8769 & 3.7481 & 2.6554 \\
        \(d=8\), per row & 2.3575 & 2.9266 & 2.9322 & 2.9809 & 2.8194 & 3.7563 & 2.6390 \\
        \(d=32\), per row & 2.3601 & 2.9213 & 2.9244 & 3.0140 & 2.8472 & 3.7441 & 2.6488 \\
        \bottomrule
    \end{tabular}
\end{table}

\begin{table}[H]
    \centering
    \scriptsize
    \caption{Change in loss from \(d=1\) on alternate-corpus sets, paired by seed
    (\(\times10^{-3}\), mean \(\pm\) 95\% \(t\) half-width, three seeds; negative is better).}
    \label{tab:ood-paired}
    \setlength{\tabcolsep}{1.5pt}
    \begin{tabular}{lccccccc}
        \toprule
        & & \multicolumn{3}{c}{Natural-language-heavy} & \multicolumn{3}{c}{Code/math/LaTeX-rich} \\
        \cmidrule(lr){3-5}\cmidrule(lr){6-8}
        Method & C4 & FPO mix & DCLM & OLM-Wiki & StarCoder & ProofPile & OLM-arXiv \\
        \midrule
        \multicolumn{8}{l}{\textit{\(2\times\)}} \\
        AdamW & \(+89.0\pm7.8\) & \(+87.9\pm53.6\) & \(+91.5\pm41.5\) & \(+104.1\pm28.2\) & \(+27.9\pm261.3\) & \(+121.8\pm182.1\) & \(+77.3\pm128.9\) \\
        Muon & \(+13.1\pm3.4\) & \(+9.0\pm14.8\) & \(+13.8\pm3.1\) & \(+14.1\pm20.3\) & \(-47.6\pm142.2\) & \(-87.2\pm82.7\) & \(-7.7\pm126.2\) \\
        MuonH & \(+6.4\pm2.3\) & \(+8.3\pm12.6\) & \(+8.8\pm19.9\) & \(+19.8\pm53.6\) & \(+9.2\pm54.7\) & \(-23.9\pm106.7\) & \(+14.8\pm200.5\) \\
        \(d=1\), no gain & \(+5.9\pm2.4\) & \(+4.9\pm17.1\) & \(+9.1\pm5.6\) & \(+20.7\pm19.3\) & \(-52.6\pm293.6\) & \(+7.0\pm83.4\) & \(+0.7\pm27.4\) \\
        \(d=8\), per row & \(+0.8\pm5.7\) & \(-2.1\pm29.4\) & \(-0.8\pm30.6\) & \(+9.1\pm49.7\) & \(-27.1\pm121.4\) & \(-6.0\pm161.0\) & \(+1.7\pm111.8\) \\
        \(d=32\), per row & \(+5.2\pm2.6\) & \(+4.7\pm12.5\) & \(+7.9\pm14.6\) & \(+4.7\pm18.2\) & \(-35.5\pm36.9\) & \(-27.3\pm118.6\) & \(+21.7\pm100.5\) \\
        \midrule
        \multicolumn{8}{l}{\textit{\(7.69\times\)}} \\
        Muon & \(+15.7\pm1.8\) & \(+13.3\pm12.9\) & \(+10.7\pm8.6\) & \(-13.2\pm28.5\) & \(+54.1\pm140.8\) & \(-19.9\pm193.4\) & \(+6.4\pm118.3\) \\
        MuonH & \(+5.9\pm3.6\) & \(+7.4\pm30.8\) & \(+6.9\pm24.5\) & \(+18.2\pm51.5\) & \(+17.9\pm141.2\) & \(-14.4\pm101.3\) & \(+44.0\pm85.7\) \\
        \(d=1\), no gain & \(+8.6\pm1.9\) & \(+2.0\pm17.6\) & \(+6.2\pm12.1\) & \(+7.0\pm26.4\) & \(-44.1\pm152.2\) & \(-38.7\pm131.5\) & \(+26.4\pm46.0\) \\
        \(d=8\), per row & \(+1.1\pm1.4\) & \(+0.0\pm52.7\) & \(+3.9\pm50.5\) & \(-58.2\pm201.5\) & \(-57.5\pm122.8\) & \(+8.2\pm39.7\) & \(-16.3\pm112.3\) \\
        \(d=32\), per row & \(+3.7\pm0.9\) & \(-5.3\pm35.7\) & \(-3.9\pm19.8\) & \(-25.0\pm1.8\) & \(-29.7\pm292.4\) & \(-4.1\pm203.9\) & \(-6.6\pm36.1\) \\
        \bottomrule
    \end{tabular}
\end{table}

\begin{table}[H]
    \centering
    \small
    \caption{Seed noise on alternate-corpus sets. Range is the spread of the three \(d=1\) seeds
    (max \(-\) min, \(\times10^{-3}\)); \(r\) is the correlation across arms between each corpus's
    seed-paired differences from \(d=1\) and those on C4 (AdamW excluded).}
    \label{tab:ood-noise}
    \begin{tabular}{lcccc}
        \toprule
        & \multicolumn{2}{c}{\(2\times\)} & \multicolumn{2}{c}{\(7.69\times\)} \\
        \cmidrule(lr){2-3}\cmidrule(lr){4-5}
        Corpus & \(d=1\) range & \(r\) & \(d=1\) range & \(r\) \\
        \midrule
        C4 & 1.6 & -- & 1.5 & -- \\
        FPO mix & 8.4 & \(+0.81\) & 6.8 & \(+0.80\) \\
        DCLM & 9.4 & \(+0.85\) & 7.9 & \(+0.70\) \\
        OLM-Wiki & 26.1 & \(+0.41\) & 20.9 & \(+0.45\) \\
        StarCoder & 78.2 & \(-0.12\) & 106.3 & \(+0.78\) \\
        ProofPile & 74.3 & \(-0.78\) & 76.1 & \(-0.65\) \\
        OLM-arXiv & 99.9 & \(+0.05\) & 59.2 & \(+0.31\) \\
        \bottomrule
    \end{tabular}
\end{table}

\section{Matrix-Gain Controls}
\label{app:gains}

\begin{table}[H]
    \centering
    \scriptsize
    \caption{Selected \(d>1\) controls at \(d=32\) unless noted; \(n\) is the
    number of training seeds and \(\Delta\) is in milli-loss relative to the
    panel's \(d=1\) reference. SPD, XXT, and XXT+diagonal use full-matrix block
    gains. Large-block and manifold-Muon results appear in the tables below.}
    \label{tab:dgreater1-attempts}
    \footnotesize
    \setlength{\tabcolsep}{3pt}
    \begin{tabular}[t]{lrrr}
        \toprule
        Realization & \(n\) & Loss & \(\Delta\) \\
        \midrule
        \multicolumn{4}{l}{360M; \(d=1=2.4621\)} \\
        Stiefel, per-block gain & 3 & 2.4686 & +6.5 \\
        SPD \(LL^\top Q\) & 3 & 2.4619 & \(-0.2\) \\
        XXT & 3 & 2.4624 & +0.2 \\
        XXT + diagonal & 3 & 2.4614 & \(-0.8\) \\
        Projected DGram & 3 & 2.4810 & +18.8 \\
        \bottomrule
    \end{tabular}\hspace{2em}
    \begin{tabular}[t]{lrrr}
        \toprule
        Realization & \(n\) & Loss & \(\Delta\) \\
        \midrule
        \multicolumn{4}{l}{1.3B; \(d=1=2.1842\)} \\
        Stiefel, per-row gain & 3 & 2.1861 & +1.9 \\
        SPD & 3 & 2.1832 & \(-1.0\) \\
        SPD, \(d=64\) & 3 & 2.1852 & +1.0 \\
        \midrule
        \multicolumn{4}{l}{3.1B; \(d=1=2.0242\)} \\
        Stiefel, per-row gain & 3 & 2.0249 & +0.8 \\
        SPD & 3 & 2.0239 & \(-0.3\) \\
        \bottomrule
    \end{tabular}
\end{table}

\begin{table}[H]
    \centering
    \small
    \caption{Complementary 360M, \(2\times\) matrix-gain study.}
    \label{tab:matrix-gain}
    \begin{tabular}{llrcc}
        \toprule
        Parameterization & Gain granularity & \(n\) & Loss &
        \(\Delta\) vs.\ \(d=1\) \\
        \midrule
        \(d=1\) & per row & 3 & 2.4621 & 0.0000 \\
        \(d=32\), scalar & per block & 3 & 2.4686 & +0.0065 \\
        \(d=32\), SPD \(LL^\top\) & full matrix & 3 & 2.4619 & \(-0.0002\) \\
        \(d=32\), XXT & full matrix & 3 & 2.4624 & +0.0002 \\
        \(d=32\), XXT + diagonal & full matrix & 3 & 2.4614 & \(-0.0008\) \\
        \(d=32\), symmetric-square + Muon & full matrix & 3 & 2.4626 & +0.0005 \\
        \bottomrule
    \end{tabular}
\end{table}

\begin{table}[H]
    \centering
    \small
    \caption{Left factors at \(d=32\) at 360M, \(2\times\) (three seeds
    each; \(\Delta\) in milli-loss from \(d=1\), 2.4621).}
    \label{tab:gain-param}
    \begin{tabular}{llcr}
        \toprule
        Left factor & Form & Loss & \(\Delta\) \\
        \midrule
        Per-row & \(\operatorname{diag}(a)\) & 2.4673 & +5.2 \\
        SPD & \(LL^\top\) & 2.4619 & \(-0.2\) \\
        SPD + diagonal & \(LL^\top+\operatorname{diag}(\lambda)\) & 2.4614 & \(-0.7\) \\
        XXT & \(XX^\top\) & 2.4624 & +0.2 \\
        Matrix exponential & \(\exp(A)\), \(A=A^\top\) & 2.4629 & +0.8 \\
        Indefinite symmetric & \(\operatorname{diag}(a)+\operatorname{Off}(M)\) & 2.4635 & +1.3 \\
        Lower triangular & \(T\) & 2.4644 & +2.3 \\
        Singular-value & \(U\operatorname{diag}(s)\), \(U^\top U=I\) & 2.4676 & +5.5 \\
        \bottomrule
    \end{tabular}
\end{table}

The symmetric factors lie within 1.3 milli-loss of \(d=1\). The
lower-triangular factor trails SPD by 2.5 milli-loss, and the singular-value
form does no better than per-row gains. For the singular-value form, frame
learning rates 0.0005 and 0.002 give the same seed-42 loss, 2.4664.

\begin{table}[H]
    \centering
    \small
    \caption{Per-seed 1.3B \(d=32\) SPD results at the independently selected
    main learning rate \(0.001\).}
    \begin{tabular}{rc}
        \toprule
        Seed & Final C4 loss \\
        \midrule
        42 & 2.1832 \\
        1234 & 2.1834 \\
        5678 & 2.1832 \\
        \midrule
        Mean \(\pm\) SD & 2.1832 \(\pm\) 0.0001 \\
        \bottomrule
    \end{tabular}
\end{table}

A five-point seed-42 bracket over
\(\{0.0005,0.0007071,0.001,0.0014142,0.002\}\) gives losses
2.2030, 2.1880, 2.1832, 2.1882, and 2.1985, respectively. The selected point
is therefore an interior optimum. The paired difference from \(d=1\) is
\(-0.98\) milli-loss; Appendix~\ref{app:gain-lr} shows that it reverses when
the gain learning rate of \(d=1\) is tuned.

\begin{table}[H]
    \centering
    \small
    \caption{Selected large-block results at 360M (three-seed means). Each
    method uses its selected main learning rate; SPD gain learning rates are
    selected separately.}
    \label{tab:large-d-full}
    \begin{tabular}{rcccc}
        \toprule
        & \multicolumn{2}{c}{\(2\times\)} & \multicolumn{2}{c}{\(4\times\)} \\
        \cmidrule(lr){2-3}\cmidrule(lr){4-5}
        \(d\) & Per-row gain & SPD gain & Per-row gain & SPD gain \\
        \midrule
        128  & 2.4814 & 2.4697 & 2.4174 & 2.4066 \\
        256  & 2.4993 & 2.4792 & 2.4345 & 2.4154 \\
        1024 & 2.6172 & 2.5465 & 2.5524 & 2.4731 \\
        \bottomrule
    \end{tabular}
\end{table}

Across both budgets, SPD recovers an increasing absolute amount of the
large-block penalty, while all six selected SPD cells remain worse than the
corresponding \(d=1\) control. The selected SPD-factor learning rates are
0.0006 at \(2\times\) for every \(d\), and 0.0012, 0.0006, and 0.0003 at
\(4\times\) for \(d=128\), 256, and 1024; at \(d=1024\), 0.0024 diverges at
both budgets.

\begin{table}[H]
    \centering
    \small
    \caption{Selected large-block results at 1.3B, \(2\times\) (three-seed
    means). Every directional-learning-rate bracket selects the interior value
    0.001. \(\Delta\) is the paired difference from \(d=1\) (2.1842).
    \(^{\dagger}\)Two converged seeds; seed 42 diverged near step 31,500. At
    learning rate 0.0005, seed 42 converges to 2.2682.}
    \label{tab:large-d-1p3b}
    \begin{tabular}{rccrr}
        \toprule
        \(d\) & Per-row gain & SPD gain & \(\Delta_{\mathrm{row}}\times10^3\) &
        \(\Delta_{\mathrm{SPD}}\times10^3\) \\
        \midrule
        128  & 2.1924 & 2.1886 & +8.2 & +4.3 \\
        256  & 2.1989 & 2.1968 & +14.7 & +12.5 \\
        1024 & 2.2411 & 2.2578\(^{\dagger}\) & +56.9 & +73.5 \\
        \bottomrule
    \end{tabular}
\end{table}

The columns \(\Delta_{\mathrm{row}}\) and \(\Delta_{\mathrm{SPD}}\) in
Table~\ref{tab:large-d-1p3b} are paired differences of the per-row-gain and
SPD-gain arms from \(d=1\) (2.1842). SPD recovers part of the large-block
penalty at \(d=128\) and 256. At \(d=1024\), SPD is both worse than the
per-row gain and unstable at the selected learning rate.

\begin{table}[H]
    \centering
    \small
    \caption{Selected 360M manifold-Muon results (three-seed means).}
    \label{tab:manifold-muon-360}
    \begin{tabular}{crlrrr}
        \toprule
        Budget & \(d\) & Gain & Manifold Muon & OrthMuon same \(d\) & \(d=1\) \\
        \midrule
        \(2\times\) & 32 & per row & 2.4977 & 2.4673 & 2.4621 \\
        \(2\times\) & 32 & SPD & 2.4947 & 2.4619 & 2.4621 \\
        \(2\times\) & 256 & per row & 2.5064 & 2.4993 & 2.4621 \\
        \(2\times\) & 1024 & per row & 2.6041 & 2.6172 & 2.4621 \\
        \midrule
        \(4\times\) & 32 & per row & 2.4280 & 2.4053 & 2.4002 \\
        \(4\times\) & 32 & SPD & 2.4260 & 2.3996 & 2.4002 \\
        \(4\times\) & 256 & per row & 2.4394 & 2.4345 & 2.4002 \\
        \(4\times\) & 1024 & per row & 2.5401 & 2.5524 & 2.4002 \\
        \bottomrule
    \end{tabular}
\end{table}

Manifold Muon improves on projected OrthMuon only at \(d=1024\), while
remaining far behind \(d=1\) throughout. At 1.3B
(Table~\ref{tab:manifold-muon-1p3b}), it trails projected OrthMuon at every
block size: by 25.7, 27.7, 8.6, and 5.8 milli-loss for per-row \(d=32\), SPD
\(d=32\), per-row \(d=256\), and per-row \(d=1024\). All four arms select
learning rate 0.0005, half the OrthMuon rate, from
\(\{0.00025,0.0005,0.001,0.002,0.004\}\).

\begin{table}[H]
    \centering
    \small
    \caption{1.3B, \(2\times\) manifold-Muon runs: three-seed means at
    learning rate 0.0005; references are three-seed means; \(\Delta\) is
    paired against \(d=1\).}
    \label{tab:manifold-muon-1p3b}
    \begin{tabular}{rlccc}
        \toprule
        \(d\) & Gain & Manifold Muon & OrthMuon same \(d\) & \(\Delta\) vs.\ \(d=1\) (\(\times10^{-3}\)) \\
        \midrule
        32 & per row & 2.2118 & 2.1861 & +27.6 \\
        32 & SPD & 2.2110 & 2.1832 & +26.8 \\
        256 & per row & 2.2076 & 2.1989 & +23.4 \\
        1024 & per row & 2.2468 & 2.2411 & +62.6 \\
        \bottomrule
    \end{tabular}
\end{table}

\section{Row Geometry of Trained Checkpoints}
\label{app:row-geometry}

We analyze the final seed-42 checkpoints of \(d=1\), per-row \(d=32\),
SPD \(d=32\), tuned Muon, and MuonH at 360M, \(2\times\). The primary \(d=1\) checkpoint uses the
default gain learning rate of Table~\ref{tab:main}; we also report a \(d=1\) checkpoint with gain
learning rate 0.0048 (Appendix~\ref{app:module-swap}). Effective weights are read from the forward
pass, so they include the gains. Every matrix is split into the
same contiguous 32-row windows in its stored-row orientation, so the \(d=1\)
Gram matrices are not trivially \(1\times1\). In this orientation every
stored row lies in \(\mathbb{R}^{1024}\), including the MLP down projection,
which is stored as a \(3072\times1024\) matrix. Statistics are pooled over
all 24 layers. For the effective rows
\(w_i\) of a window, the cosine is
\(C_{ij}=\langle w_i,w_j\rangle/(\lVert w_i\rVert_2\lVert w_j\rVert_2)\).
Each arm contributes 9{,}984 windows and about 4.95M row pairs. For
independent isotropic rows of width 1024, the expected absolute cosine is
\(\sqrt{2/(1024\pi)}\approx0.025\), and blocks of 32 random unit rows
have median cosine-matrix largest eigenvalue 1.34 and median condition number
1.38 (4{,}000 simulated blocks). Stable rank is
\(\lVert W\rVert_F^2/\lVert W\rVert_2^2\) of a \(32\)-row block.

\begin{table}[H]
    \centering
    \small
    \caption{Geometry of effective 32-row blocks pooled over all matrices.
    Mean and p95 are over row pairs; the other columns are medians over
    blocks.}
    \label{tab:row-geometry-all}
    \begin{tabular}{lccccc}
        \toprule
        Arm & Mean \(|C_{ij}|\) & p95 \(|C_{ij}|\) & \(\lambda_{\max}(C)\) & Stable rank & Condition number \\
        \midrule
        Random unit rows & 0.025 & -- & 1.34 & 23.9 & 1.38 \\
        \midrule
        Muon & 0.032 & 0.080 & 1.385 & 20.4 & 1.61 \\
        MuonH & 0.034 & 0.085 & 1.385 & 20.5 & 1.61 \\
        \(d=1\) & 0.030 & 0.076 & 1.380 & 21.6 & 1.54 \\
        \(d=1\), gain LR 0.0048 & 0.030 & 0.076 & 1.377 & 19.2 & 1.74 \\
        Per-row \(d=32\) & \(<10^{-8}\) & \(<10^{-7}\) & 1.000 & 25.3 & 1.30 \\
        SPD \(d=32\) & 0.052 & 0.142 & 1.777 & 16.9 & 1.95 \\
        \bottomrule
    \end{tabular}
\end{table}

\begin{table}[H]
    \centering
    \small
    \caption{Absolute cosine between effective rows in the same 32-row window.
    Per-row \(d=32\) is exactly zero (below \(10^{-7}\)) in every matrix type.}
    \begin{tabular}{lcccccc}
        \toprule
        & \multicolumn{2}{c}{Muon} & \multicolumn{2}{c}{\(d=1\)} & \multicolumn{2}{c}{SPD \(d=32\)} \\
        \cmidrule(lr){2-3}\cmidrule(lr){4-5}\cmidrule(lr){6-7}
        Matrix & Mean & p95 & Mean & p95 & Mean & p95 \\
        \midrule
        Query & 0.052 & 0.162 & 0.047 & 0.138 & 0.105 & 0.328 \\
        Key & 0.055 & 0.159 & 0.047 & 0.134 & 0.111 & 0.324 \\
        Value & 0.045 & 0.149 & 0.031 & 0.086 & 0.048 & 0.132 \\
        Attention output & 0.028 & 0.071 & 0.028 & 0.068 & 0.047 & 0.116 \\
        MLP gate & 0.026 & 0.065 & 0.026 & 0.065 & 0.044 & 0.112 \\
        MLP up & 0.026 & 0.063 & 0.026 & 0.063 & 0.039 & 0.097 \\
        MLP down & 0.027 & 0.066 & 0.026 & 0.066 & 0.038 & 0.096 \\
        \midrule
        All & 0.032 & 0.080 & 0.030 & 0.076 & 0.052 & 0.142 \\
        \bottomrule
    \end{tabular}
\end{table}

Trained \(d=1\) rows stay close to the random scale in the MLP but become
more coherent in query and key matrices. Unconstrained Muon rows are slightly
more coherent than \(d=1\) rows in query, key, and value matrices and match
them elsewhere;
MuonH is within 0.002 of Muon in pooled mean absolute cosine. Pooled over all matrices, the
\(d=1\) directional factor is close to the random level: its median block
condition number is 1.42, against 1.38 for random unit rows. The per-row \(d=32\) constraint
removes this coherence exactly, and the SPD gain restores and amplifies it
while its directional factor \(Q\) remains orthogonal; the increase is largest
in query and key matrices. Condition numbers show the same pattern. The median
condition number of an effective 32-row window is 2.9--5.2 for \(d=1\) key
and query matrices (3.1 and 2.9 at the default gain learning rate, 5.2 and 4.8 at 0.0048), 2.5 and
2.3 for per-row \(d=32\), and 8.0 and 7.3 for SPD \(d=32\). Blockwise orthogonality does not make
whole matrices well conditioned: the median condition number of the \(1024\times1024\) key and query
matrices is \(7\times10^3\)--\(3\times10^4\) for \(d=1\) (\(6.8\times10^3\) and \(1.0\times10^4\) at
the default gain learning rate, \(2.5\times10^4\) and \(2.8\times10^4\) at 0.0048) and about
\(7\times10^3\) for per-row \(d=32\). Per-row \(d=32\) improves block conditioning by
construction and has the highest stable rank
(Table~\ref{tab:row-geometry-all}), yet it has higher loss than \(d=1\).
The gain learning rate of \(d=1\) alone moves its median block condition number from 1.54 to 1.74
and its stable rank from 21.6 to 19.2 while changing the loss by less than 2 milli-loss. Better
conditioning and higher stable rank therefore do not translate into lower validation loss.

\begin{table}[H]
    \centering
    \small
    \caption{Geometry of effective 128- and 1024-row windows pooled over all
    matrices. Mean is over row pairs; the other columns are medians over
    windows. Pooled condition numbers of 1024-row windows mix matrix types and
    are omitted.}
    \label{tab:row-geometry-windows}
    \begin{tabular}{llcccc}
        \toprule
        Window & Arm & Mean \(|C_{ij}|\) & \(\lambda_{\max}(C)\) & Stable rank & Condition number \\
        \midrule
        128 & Random unit rows & 0.025 & 1.80 & 71.1 & 2.04 \\
            & Muon & 0.030 & 2.00 & 58.5 & 2.47 \\
            & \(d=1\) & 0.028 & 1.96 & 61.7 & 2.35 \\
            & Per-row \(d=32\) & 0.021 & 1.72 & 69.0 & 2.14 \\
            & SPD \(d=32\) & 0.035 & 2.46 & 48.3 & 2.88 \\
        \midrule
        1024 & Random unit rows & 0.025 & 3.94 & 260 & -- \\
             & Muon & 0.028 & 7.16 & 146.0 & -- \\
             & \(d=1\) & 0.027 & 6.71 & 152.5 & -- \\
        \bottomrule
    \end{tabular}
\end{table}

Larger windows give the same ordering (Table~\ref{tab:row-geometry-windows}).
In 128-row windows, 1{,}984 of the 8{,}128 row pairs lie in the same
\(d=32\) block, so the across-block pairs of per-row \(d=32\) average
0.027, close to 0.028 for \(d=1\) rows; block orthogonality does not propagate
beyond the block. In 1024-row windows, \(d=1\) and Muon have nearly the same
spectrum, with largest cosine-matrix eigenvalues 6.7 and 7.2 against 3.9 for
random rows and stable ranks 153 and 146 against 260. This large-scale
concentration therefore arises without any constraint, and the sphere
constraint does not change it.

\section{Projected DGram}
\label{app:dgram}

\begin{table}[H]
    \centering
    \small
    \caption{Projected DGram \(d=32\) LR--WD sweep at 360M, \(2\times\). Entries are two-seed means.}
    \begin{tabular}{rccc}
        \toprule
        LR & WD 0.1 & WD 0.3 & WD 1.0 \\
        \midrule
        0.002 & 2.5140 & 2.4879 & 2.4907 \\
        0.004 & 2.5010 & \textbf{2.4808} & 2.5055 \\
        0.008 & 2.8647 & 2.5073 & 2.5461 \\
        \bottomrule
    \end{tabular}
\end{table}

\begin{table}[H]
    \centering
    \small
    \caption{Projected DGram block-size screen without weight decay:
    three-seed means at the seed-42 pick.}
    \begin{tabular}{rcc}
        \toprule
        Block size & Selected LR & Final loss \\
        \midrule
        32 & 0.002 & 2.5445 \\
        128 & 0.002 & 2.5581 \\
        1024 & 0.004 & 2.6881 \\
        \bottomrule
    \end{tabular}
\end{table}

At \(d=1024\), 0.008 diverges.

\section{Module-Wise Substitution}
\label{app:module-swap}

Table~\ref{tab:module-swap} tests module assignment at 360M, \(2\times\), with
seeds 42, 1234, and 5678. One module type, attention (\(q,k,v,o\)) or MLP (gate, up, down), stays
under \(d=1\) with per-row gains, directional learning rate 0.002, weight decay
0.1, and gain learning rate 0.0048, four times the shared AdamW learning rate of
0.0012, which is unchanged. The other module type is
replaced. The reference trains every matrix with these \(d=1\) settings and
reaches 2.4604, the three-seed mean. Replacement settings are the tuned settings of each method:
Muon uses learning rate 0.004 and weight decay 0.3; MuonH uses 0.0125 and no
weight decay; per-row and SPD \(d=32\) use directional learning rate 0.002,
weight decay 0.1, and gain learning rate 0.0012, with no weight decay on the
SPD factor; projected DGram uses \(d=32\), learning rate 0.004, and weight
decay 0.3. The replaced \(d=32\) modules thus use a smaller gain learning rate
than the fixed \(d=1\) modules and the reference.

\begin{table}[H]
    \centering
    \small
    \caption{Module-wise substitution at 360M, \(2\times\): three-seed mean
    final C4 loss and paired change from the all-\(d=1\) reference 2.4604
    (\(\times10^{-3}\); negative is better).}
    \label{tab:module-swap}
    \begin{tabular}{lcccc}
        \toprule
        & \multicolumn{2}{c}{MLP replaced} & \multicolumn{2}{c}{Attention replaced} \\
        \cmidrule(lr){2-3}\cmidrule(lr){4-5}
        Replacement & Loss & \(\Delta\) & Loss & \(\Delta\) \\
        \midrule
        Muon & 2.4712 & +10.8 & 2.4679 & +7.5 \\
        MuonH & 2.4617 & +1.4 & 2.4679 & +7.5 \\
        Per-row \(d=32\) & 2.4630 & +2.6 & 2.4635 & +3.2 \\
        SPD \(d=32\) & 2.4619 & +1.6 & 2.4582 & \(-2.1\) \\
        Projected DGram & 2.4730 & +12.6 & 2.4707 & +10.3 \\
        \bottomrule
    \end{tabular}
\end{table}

Projected DGram is the worst replacement in either module, and it is worst
in MLP matrices, where \citet{yoshihara2026manifold} assign it. Per-row
\(d=32\) is worse than \(d=1\) in both module types. MuonH is close to \(d=1\)
in MLP matrices but not in attention, whereas Muon trails in both. SPD
\(d=32\) in attention is the only replacement below the reference, on every
seed (95\% interval \([-3.99,-0.28]\)); Section~\ref{sec:qk-spd} narrows this
to queries and keys. Each replacement prefers the same module type in all
three seeds: MuonH and per-row \(d=32\) do better in MLP matrices, and Muon,
SPD \(d=32\), and projected DGram do better in attention matrices.

\section{SPD Factors on Single Projection Types}
\label{app:qk-spd}

Section~\ref{sec:qk-spd} gives one projection type SPD factors with block size
64 and keeps every other matrix at \(d=1\). The target projection uses
OrthMuon with an SPD factor, trained at factor learning rate 0.0012 with no
weight decay; every other matrix uses \(d=1\) with gain learning rate 0.0048.
The reference trains every matrix with \(d=1\) at gain learning rate 0.0048.
The directional learning rate is 0.002 at 360M, 0.001 at 1.3B, and 0.0005 at
3.1B. Table~\ref{tab:qk-spd-full} gives per-seed losses and paired
differences from the reference of the same model and budget.

\begin{table}[H]
    \centering
    \small
    \caption{SPD factors with block size 64 on one projection type: per-seed
    final C4 loss, three-seed mean, and paired change from the all-\(d=1\)
    reference at gain learning rate 0.0048 (\(\times10^{-3}\), mean \(\pm\)
    95\% half-width; negative is better).}
    \label{tab:qk-spd-full}
    \setlength{\tabcolsep}{4pt}
    \begin{tabular}{lccccr}
        \toprule
        SPD factor on & Seed 42 & Seed 1234 & Seed 5678 & Mean & \(\Delta\) \\
        \midrule
        \multicolumn{6}{l}{\emph{360M, \(2\times\)}} \\
        None (reference) & 2.4608 & 2.4595 & 2.4608 & 2.4604 & -- \\
        Query and key & 2.4556 & 2.4557 & 2.4561 & 2.4558 & \(-4.57\pm1.71\) \\
        Query & 2.4580 & 2.4583 & 2.4583 & 2.4582 & \(-2.17\pm2.01\) \\
        Key & 2.4581 & 2.4580 & 2.4582 & 2.4581 & \(-2.28\pm1.52\) \\
        Value & 2.4609 & 2.4607 & 2.4623 & 2.4613 & \(+0.95\pm1.84\) \\
        Output & 2.4603 & 2.4602 & 2.4609 & 2.4604 & \(+0.07\pm1.47\) \\
        MLP gate & 2.4613 & 2.4622 & 2.4618 & 2.4617 & \(+1.38\pm2.80\) \\
        MLP up & 2.4602 & 2.4605 & 2.4613 & 2.4607 & \(+0.29\pm1.98\) \\
        MLP down & 2.4632 & 2.4614 & 2.4626 & 2.4624 & \(+2.08\pm0.79\) \\
        \midrule
        \multicolumn{6}{l}{\emph{360M, \(4\times\)}} \\
        None (reference) & 2.3985 & 2.3987 & 2.3998 & 2.3990 & -- \\
        Query and key & 2.3965 & 2.3949 & 2.3957 & 2.3957 & \(-3.28\pm2.83\) \\
        Query & 2.3968 & 2.3965 & 2.3971 & 2.3968 & \(-2.20\pm1.21\) \\
        Key & 2.3970 & 2.3978 & 2.3962 & 2.3970 & \(-2.01\pm3.59\) \\
        Value & 2.3983 & 2.3999 & 2.4001 & 2.3994 & \(+0.44\pm1.78\) \\
        Output & 2.4000 & 2.4002 & 2.3989 & 2.3997 & \(+0.69\pm3.41\) \\
        MLP gate & 2.3995 & 2.4012 & 2.4002 & 2.4003 & \(+1.25\pm2.77\) \\
        MLP up & 2.3991 & 2.4010 & 2.4005 & 2.4002 & \(+1.18\pm2.44\) \\
        MLP down & 2.4015 & 2.3984 & 2.4002 & 2.4000 & \(+1.01\pm4.26\) \\
        \midrule
        \multicolumn{6}{l}{\emph{360M, \(7.69\times\)}} \\
        None (reference) & 2.3553 & 2.3563 & 2.3558 & 2.3558 & -- \\
        Query and key & 2.3552 & 2.3517 & 2.3522 & 2.3531 & \(-2.72\pm5.91\) \\
        \midrule
        \multicolumn{6}{l}{\emph{1.3B, \(2\times\)}} \\
        None (reference) & 2.1824 & 2.1836 & 2.1821 & 2.1827 & -- \\
        Query and key & 2.1814 & 2.1818 & 2.1816 & 2.1816 & \(-1.10\pm1.59\) \\
        Query & 2.1815 & 2.1816 & 2.1816 & 2.1816 & \(-1.11\pm1.76\) \\
        Key & 2.1823 & 2.1821 & 2.1825 & 2.1823 & \(-0.38\pm2.39\) \\
        Value & 2.1831 & 2.1839 & 2.1839 & 2.1836 & \(+0.93\pm1.85\) \\
        Output & 2.1840 & 2.1841 & 2.1838 & 2.1839 & \(+1.26\pm1.57\) \\
        MLP gate & 2.1837 & 2.1839 & 2.1839 & 2.1839 & \(+1.17\pm1.81\) \\
        MLP up & 2.1845 & 2.1838 & 2.1837 & 2.1840 & \(+1.35\pm2.45\) \\
        MLP down & 2.1840 & 2.1844 & 2.1841 & 2.1842 & \(+1.48\pm1.46\) \\
        \midrule
        \multicolumn{6}{l}{\emph{3.1B, \(2\times\), query and key}} \\
        None (reference) & 2.0242 & 2.0235 & 2.0240 & 2.0239 & -- \\
        Block size 64 & 2.0238 & 2.0224 & 2.0233 & 2.0232 & \(-0.71\pm1.00\) \\
        Block size 128 & 2.0245 & 2.0245 & 2.0248 & 2.0246 & \(+0.71\pm0.87\) \\
        \bottomrule
    \end{tabular}
\end{table}

\paragraph{Training length and factor learning rate.}
At 360M, the query-and-key gain shrinks from \(-4.57\pm1.71\) at \(2\times\)
to \(-3.28\pm2.83\) at \(4\times\) and \(-2.72\pm5.91\) at \(7.69\times\), with
per-seed differences \(-0.04\), \(-4.58\), and \(-3.55\) at \(7.69\times\).
At 360M, \(2\times\), the query-and-key factor learning rate 0.0012 is
interior on seed 42: 0.0006, 0.0012, and 0.0024 give 2.4578, 2.4556, and
2.4559.

\paragraph{3.1B.}
The 3.1B model has head dimension 128, with 22 query and 11 key heads per
layer. Its reference, 2.0239, is \(-0.27\pm0.34\) milli-loss against
\(d=1\) at the default gain learning rate. With block size 64, SPD factors on
queries and keys change loss by \(-0.71\pm1.00\), with per-seed differences
\(-0.33\), \(-1.13\), and \(-0.66\). With block size 128, one factor per head,
the change is \(+0.71\pm0.87\), with per-seed differences \(+0.32\),
\(+0.98\), and \(+0.84\). Block size 128 minus block size 64 is
\(+1.42\pm1.83\).

\paragraph{Attention logits.}
We measure the 3.1B finals on 4 sequences of 4096 C4-val tokens. The max
logit is the largest \(q\cdot k/\sqrt{128}\) after RoPE; the median head max
is the median over heads of each head's largest logit; and
\(\lVert W_q\rVert\cdot\lVert W_k\rVert\) is the product of a head's query and
key spectral norms, gains included. Table~\ref{tab:qk-logits} averages each
statistic over the three seeds. The per-run max logits span 56--222 for the
reference, 42--214 for block size 64, and 307--5170 for block size 128,
whereas the median head max stays near 10 in every run. Per-row gains bound
the norm of each head through its gains, whereas \(S=LL^\top\) places no cap
on its eigenvalues. The higher gain learning rate alone also raises the bound:
at the default gain learning rate, the seed-42 \(d=1\) run has max
\(\lVert W_q\rVert\cdot\lVert W_k\rVert\) 46.0 and max logit 41.3. Max logits
are extreme-value statistics from one measurement batch.

\begin{table}[H]
    \centering
    \small
    \caption{Attention logits of the 3.1B query-and-key finals, averaged
    over three seeds; the last column gives the range of the per-run max
    logit.}
    \label{tab:qk-logits}
    \begin{tabular}{lcccc}
        \toprule
        Arm & Max logit & Median head max & Max \(\lVert W_q\rVert\cdot\lVert W_k\rVert\) & Max-logit range \\
        \midrule
        Reference & 112.7 & 9.5 & 78.3 & 56--222 \\
        Block size 64 & 110.6 & 10.0 & 144.8 & 42--214 \\
        Block size 128 & 1976.0 & 10.7 & 402.2 & 307--5170 \\
        \bottomrule
    \end{tabular}
\end{table}

\paragraph{Relation to module-wise substitution.}
SPD \(d=32\) on all four attention projections changes loss by
\(-2.13\pm1.85\) (Appendix~\ref{app:module-swap}). SPD factors on queries and
keys at block size 64 improve on it by \(2.44\pm0.87\), but the block size
also differs (32 against 64).

\paragraph{Cost and grouped-query attention.}
Each 64-row block carries an SPD factor with \(64\cdot65/2=2080\) parameters
in place of 64 row gains. On queries and keys, this amounts to about 1.6M
factor parameters at 360M (0.44\% of the model), 2.4M at 1.3B (0.19\%), and
4.4M at 3.1B with block size 64 (0.14\%). The 1.3B and 3.1B models share
each key head between two query heads, and at 1.3B the query factor alone
matches query and key factors together.

\section{Gain Learning Rate}
\label{app:gain-lr}

The protocol trains every gain and SPD factor with AdamW at learning rate
0.0012. Table~\ref{tab:gain-lr} raises the gain learning rate of \(d=1\) to
0.0048, leaving every other learning rate unchanged.

\begin{table}[H]
    \centering
    \small
    \caption{Paired change in final C4 loss when the gain learning rate of
    \(d=1\) rises from 0.0012 to 0.0048 (\(\times10^{-3}\); negative is
    better).}
    \label{tab:gain-lr}
    \begin{tabular}{lcc}
        \toprule
        Model, budget & Mean \(\pm\) 95\% half-width & Seeds 42 / 1234 / 5678 \\
        \midrule
        360M, \(2\times\) & \(-1.78\pm3.92\) & \(-0.75\) / \(-3.59\) / \(-0.99\) \\
        360M, \(4\times\) & \(-1.18\pm0.91\) & \(-1.39\) / \(-1.39\) / \(-0.75\) \\
        360M, \(7.69\times\) & \(-0.62\pm1.76\) & \(-0.45\) / \(-0.02\) / \(-1.41\) \\
        1.3B, \(2\times\) & \(-1.53\pm0.46\) & \(-1.68\) / \(-1.33\) / \(-1.59\) \\
        3.1B, \(2\times\) & \(-0.27\pm0.34\) & \(-0.29\) / \(-0.40\) / \(-0.13\) \\
        \bottomrule
    \end{tabular}
\end{table}

\paragraph{Brackets.}
On seed 42, the 1.3B \(d=1\) gain learning rates 0.0012, 0.0024, 0.0048, and
0.0096 give 2.1840, 2.1829, 2.1824, and 2.1827, and the 360M \(d=1\) gain
learning rates 0.0012, 0.0024, and 0.0048 give 2.4616, 2.4607, and 2.4608.
The \(d=32\) SPD factor learning rate is best at the default: at 1.3B, seed
42, 0.0006, 0.0012, and 0.0024 give 2.1840, 2.1832, and 2.1839; at 360M,
seed 42, 0.0006, 0.0012, 0.0024, and 0.0048 give 2.4623, 2.4614, 2.4609, and
2.4635, and over three seeds 0.0024 minus 0.0012 is \(+0.06\pm1.52\).

\paragraph{SPD against the gain-tuned \(d=1\).}
Full-network SPD \(d=32\) minus \(d=1\) at gain learning rate 0.0048 is
\(+1.54\pm2.27\) at 360M, \(2\times\); \(+0.57\pm0.56\) at 360M, \(4\times\)
(per seed \(+0.31\), \(+0.73\), \(+0.68\)); and \(+0.56\pm1.67\) at 1.3B,
\(2\times\) (per seed \(+0.80\), \(-0.20\), \(+1.07\)).

\paragraph{Per-row ladder.}
The per-row ladder gap does not depend on the gain learning rate. At 360M,
seed 42, \(d=32\) minus \(d=1\) is \(+4.41\), \(+5.13\), and \(+4.24\) at gain
learning rates 0.0012, 0.0024, and 0.0048. At 1.3B, seed 42, per-row
\(d=32\) at 0.0048 reaches 2.1848, against 2.1824 for \(d=1\) at 0.0048, a gap
of \(+2.5\); at 0.0012 the gap is \(+2.3\).

Muon and MuonH have no gain, so tuning it only widens the lead of \(d=1\).
The main tables keep the 0.0012 protocol.

\section{Additional Optimizer and Coordinate Controls}
\label{app:optimizer-controls}

\begin{table}[H]
    \centering
    \small
    \caption{Selected 360M, \(2\times\), \(d=1\) controls.}
    \label{tab:d1-controls}
    \begin{tabular}{lcc}
        \toprule
        Variant & \(n\) & Final loss \\
        \midrule
        Default OrthMuon & 3 & 2.4621 \\
        Direct row normalization & 3 & 2.4622 \\
        Exponential gain & 3 & 2.4625 \\
        Tangent gradient before momentum & 3 & 2.4637 \\
        Tangent + transported momentum & 3 & 2.4628 \\
        Block-Frobenius tangent \(d=32\) + row gain & 3 & 2.4641 \\
        Forward-normalized Muon & 3 & 2.4697 \\
        Polar-split update only & 3 & 2.5680 \\
        \bottomrule
    \end{tabular}
\end{table}

Forward-normalized Muon uses LR 0.004 and WD 0.3, the seed-42 optimum of its
grid; the previously reported point, LR 0.002 and WD 1.0, gives 2.4716. The
block-Frobenius tangent control constrains only the norm of each 32-row block,
not pairwise angles; its learning rate 0.016 is interior to
\(\{0.008,0.016,0.032\}\) (2.4723, 2.4648, 2.5352 on seed 42). Polar-split
uses scale 0.6, selected from \(\{0.15,0.3,0.6,1.0\}\).
The polar-split control applies the polar decomposition to the Muon update
without maintaining a constrained weight state.

\section{Diffusion Transformer on ImageNet}
\label{app:dit}

\paragraph{Setup.}
The model is SiT-B/2, the DiT-B/2 architecture with width 768, 12 layers, and
12 heads of dimension 64; it has 130.5M parameters, of which 130.3M are
trainable. The query, key, value, output, and MLP matrices (84.9M parameters,
65\% of the total) receive the orthogonal parameterization; the adaLN
modulation (42.5M), the embedders, and the final layer stay on AdamW in every
arm. Training uses ImageNet at \(256\times256\), pre-encoded once with the
SD-VAE ft-EMA autoencoder; each step draws a horizontal flip and a fresh
posterior sample, which matches on-the-fly encoding in distribution. The global
batch is 256 images (65{,}536 tokens per step). Training runs 500K steps with
1\% linear warmup, a constant learning rate to 400K, and linear decay to zero
over the last 100K steps, under bf16 autocast with fp32 weights,
gradient-norm clipping at 1.0, and an EMA decay of 0.9999. Weight decay is
decoupled and is not applied to 1-D parameters or the class-embedding table.
Query--key normalization applies a per-head RMSNorm to queries and keys, with
learnable scales at the main learning rate and no decay. The Adam family uses
\(\beta=(0.9,0.999)\); the Muon family uses momentum 0.95, Nesterov, and five
Newton--Schulz steps. OrthAdam is the OrthMuon parameterization (unit-row
directions with per-row gains) with an Adam update. Stiefel factors and gains
receive no decay. Validation loss is measured on EMA weights over 10K
validation images at 16 fixed time values with fixed noise. FID-50K comes from
the official ADM evaluator with the SDE sampler (250 steps), no guidance, and
sampling seed 0; sFID, IS, precision, and recall come from the same evaluator.
Relative to upstream SiT, the recipe changes the latent pipeline
(pre-encoded), the schedule (warmup and decay), the precision (bf16), and adds
clipping, weight decay, and query--key normalization. The upstream recipe
reproduces FID 34.4 at 400K, against about 33 reported. Each run costs about
\(1.7\times10^{19}\) FLOPs.

\paragraph{Learning-rate selection.}
Learning rates are selected under query--key normalization by 500K EMA
validation loss, with FID breaking ties within 0.0003
(Table~\ref{tab:dit-lr}). AdamW uses weight decay 0.03, MuonW uses Muon
weight decay 0.01, and OrthMuon uses gain learning rate 0.0012. AdamW at 0.0004 ties 0.0002 on validation loss and
has lower FID (25.67 against 25.80); MuonW at 0.001 ties 0.0005 and has lower
FID (21.62 against 22.14). AdamW weight decay 0.1 is worse than 0.03, and MuonW
Muon weight decay 0.003 is worse than 0.01. The query--key-only \(d=64\) arms
run at the \(d=1\) learning rate of their family, so the \(d=64\) minus \(d=1\)
comparison is paired.

\begin{table}[H]
    \centering
    \small
    \caption{SiT-B/2 learning-rate brackets under query--key normalization
    (500K EMA validation loss): every selected learning rate is interior.}
    \label{tab:dit-lr}
    \begin{tabular}{lll}
        \toprule
        Family & Learning rate (validation loss) & Selected \\
        \midrule
        AdamW & 0.0001 (0.7578), 0.0002 (0.7562), 0.0004 (0.7561), 0.0008 (0.7575) & 0.0004 \\
        MuonW & 0.0005 (0.7535), 0.001 (0.7532), 0.002 (0.7544), 0.004 (0.7573) & 0.001 \\
        OrthAdam \(d=1\) & 0.0001 (0.7576), 0.0002 (0.7554), 0.0004 (0.7562) & 0.0002 \\
        OrthMuon \(d=1\) & 0.000125 (0.7537), 0.00025 (0.7525), 0.0005 (0.7540) & 0.00025 \\
        \bottomrule
    \end{tabular}
\end{table}

\paragraph{Main results.}
Table~\ref{tab:dit-main} reports means over five seeds (42--46) and
Table~\ref{tab:dit-paired} the seed-paired differences. All five seeds favor
the \(d=1\) arm over its baseline on all three metrics in both families.
Pooled within-arm seed standard deviations are 0.00018 for validation loss,
0.21 for FID, and 0.55 for IS. sFID, precision, and recall move with FID; for
example, precision is 0.574--0.581 for AdamW and 0.597--0.606 for OrthMuon
\(d=1\).

\begin{table}[H]
    \centering
    \small
    \caption{SiT-B/2 on ImageNet 256 with query--key normalization, means over
    five seeds: \(d=1\) improves on both baselines, and 64-row query/key blocks
    do not improve on \(d=1\).}
    \label{tab:dit-main}
    \begin{tabular}{lccc}
        \toprule
        Method & 500K EMA validation loss & FID-50K & IS \\
        \midrule
        AdamW & 0.75623 & 25.50 & 56.69 \\
        OrthAdam \(d=1\) & 0.75574 & \textbf{24.60} & 59.51 \\
        OrthAdam query--key-only \(d=64\) & 0.75607 & 25.28 & 58.67 \\
        \midrule
        MuonW & 0.75321 & 21.55 & 65.35 \\
        OrthMuon \(d=1\) & \textbf{0.75242} & \textbf{20.81} & 67.54 \\
        OrthMuon query--key-only \(d=64\) & 0.75264 & 21.06 & 67.01 \\
        \bottomrule
    \end{tabular}
\end{table}

\begin{table}[H]
    \centering
    \small
    \caption{Seed-paired SiT-B/2 differences (mean \(\pm\) 95\% \(t\)
    half-width, 4 df): \(d=1\) beats both baselines on every metric, and
    64-row query/key blocks do not beat \(d=1\).}
    \label{tab:dit-paired}
    \begin{tabular}{lccc}
        \toprule
        Difference & \(\Delta\) validation loss & \(\Delta\) FID & \(\Delta\) IS \\
        \midrule
        OrthAdam \(d=1\) \(-\) AdamW & \(-0.00049\pm0.00030\) & \(-0.90\pm0.43\) & \(+2.82\pm0.81\) \\
        OrthMuon \(d=1\) \(-\) MuonW & \(-0.00078\pm0.00021\) & \(-0.74\pm0.43\) & \(+2.19\pm1.22\) \\
        OrthAdam query--key \(d=64\) \(-\) \(d=1\) & \(+0.00032\pm0.00041\) & \(+0.68\pm0.44\) & \(-0.84\pm0.68\) \\
        OrthMuon query--key \(d=64\) \(-\) \(d=1\) & \(+0.00021\pm0.00012\) & \(+0.25\pm0.30\) & \(-0.53\pm1.04\) \\
        \bottomrule
    \end{tabular}
\end{table}

\paragraph{SPD on queries and keys.}
With SPD factors on 64-row query and key blocks (factor learning rate 0.0012,
no decay), OrthMuon reaches validation loss 0.75243 and FID 20.60 on one seed,
against 0.75249 and 20.61 for OrthMuon \(d=1\) on the same seed. This is a tie,
inside the five-seed ranges of \(d=1\) (0.75235--0.75249 and 20.59--21.09).

\paragraph{Stability without query--key normalization.}
Table~\ref{tab:dit-stability} repeats the recipe without query--key
normalization. A run counts as diverged if a log interval after 20K steps has
at least 20 clipped steps or a pre-clip gradient norm above 10. The query--key
scale is the largest singular value of the per-head product
\(W_q^{h\top}W_k^h\), maximized over heads and layers at 480K steps; it is a
weight-space scale, not a logit bound, since without normalization the logit
also depends on the input norms. Every divergence came late in the decay,
after normal gradients: AdamW on seed 44 at 486.6K steps (learning rate
\(2.7\times10^{-5}\)), OrthAdam \(d=1\) on seed 44 at 490.7K, and OrthMuon
\(d=1\) at learning rate 0.000125 at 457.6K. Scalar-gain \(d>1\) blocks keep the scale at most 9
and never diverged, including on seed 44, where AdamW and \(d=1\) diverged; an
SPD factor removes that bound (73). A large scale alone does not predict
divergence: OrthAdam \(d=1\) reached 140 on seed 43 without diverging, and
MuonW 457. With query--key normalization, none of 41 runs diverged, including
seed 44 in every arm and OrthMuon \(d=1\) at learning rate 0.000125, the
setting that diverged without normalization. Without normalization, the
non-diverged runs also reach higher FID: 25.86 and 26.15 for OrthAdam \(d=1\)
against 24.60 with normalization, and 25.83--26.20 for query--key-only
\(d=64\).

\begin{table}[H]
    \centering
    \small
    \caption{SiT-B/2 without query--key normalization: scalar-gain \(d>1\)
    blocks bound the query--key scale and never diverge.}
    \label{tab:dit-stability}
    \begin{tabular}{lccc}
        \toprule
        Arm & Seeds & Diverged & Query--key scale at 480K \\
        \midrule
        AdamW & 3 & 1 & 59--629 \\
        OrthAdam \(d=1\) & 3 & 1 & 63--140 \\
        OrthAdam query--key-only \(d=64\) & 3 & 0 & 3.5--6.0 \\
        OrthAdam \(d=32\) & 2 & 0 & 6.2--8.8 \\
        OrthAdam \(d=32\) SPD & 1 & 0 & 73 \\
        MuonW & 1 & 0 & 457 \\
        OrthMuon \(d=1\), LR 0.00025 & 1 & 0 & 98 \\
        OrthMuon \(d=1\), LR 0.000125 & 1 & 1 & 121 \\
        OrthMuon query--key-only \(d=64\), LR 0.0005 & 1 & 0 & 8.6 \\
        \bottomrule
    \end{tabular}
\end{table}

\paragraph{Code and license.}
The diffusion and sampling code (\texttt{transport/}) and the SiT model are
upstream SiT, released under the MIT license. The model file adds options that
are off by default and matches upstream exactly when they are off. The
training loop, data pipeline, and evaluation scripts are new, and the
optimizers share the language-model code base.

\paragraph{Caveats.}
The query--key-only arms have no learning-rate bracket of their own. AdamW
weight decay 0.01 and MuonW Muon weight decay 0.03 were not run under
query--key normalization. The evaluator is a vendored copy of the ADM
evaluator with two marked edits, the Inception path and one TF2 shape line.

\section{Downstream Fine-Tuning}
\label{app:sft}

\paragraph{Setup.}
The bases are the 24 final 3.1B \(2\times\) checkpoints (eight arms, three
seeds each). Each is fully fine-tuned on Commonsense170K (170{,}300 training
rows, with 120 held out for validation) using the Alpaca prompt of StelLA and a
loss on the whole sequence. Two settings differ from StelLA: no BOS token is
added, because pretraining never saw one, and the cutoff is 2048 tokens,
because 256 truncated 63.7\% of HellaSwag rows. Training runs three epochs
(31{,}932 steps) at batch size 16 with 500 warmup steps and linear decay, no
weight decay, gradient clipping at 1.0, and bf16. The reported checkpoint
minimizes validation loss, which is the end of epoch 2 for every run; the
final checkpoint agrees. Evaluation uses the corrected evaluator of StelLA on
BoolQ, PIQA, SIQA, HellaSwag, WinoGrande, ARC-e, ARC-c, and OBQA (22{,}419
prompts), and macro accuracy is the equal-weight mean over the eight tasks.
Fine-tuning runs in two columns. Under ``own optimizer'', each base keeps its
pretraining optimizer and parameterization with every learning rate scaled by
one factor. Under ``AdamW'', each base is merged into dense effective weights
and fine-tuned with one AdamW recipe (\(\beta=(0.9,0.999)\)). Learning rates
are selected on seed 42 by minimum validation loss, and every pick is interior:
under the own optimizer, scale \(1/100\) for seven arms and \(1/300\) for
Muon, over \(\{1/10,\dots,1/1000\}\); under AdamW, \(1.2\times10^{-5}\) for
seven arms and \(4\times10^{-6}\) for Muon. The fine-tuning seed equals the
pretraining seed, so arms pair by seed. The control is \(d=1\) (gain learning
rate 0.0012) for per-row \(d=32\), SPD \(d=32\), MuonH, and Muon, and all-\(d=1\)
with gain learning rate 0.0048 for the query--key arms.

\begin{table}[H]
    \centering
    \small
    \caption{Eight-task macro accuracy after fine-tuning the 3.1B models on
    Commonsense170K (percent, means over three seeds; last column paired, 95\%
    half-width): AdamW fine-tuning on merged weights raises every model by
    1.5 to 3.1 points.}
    \begin{tabular}{lccc}
        \toprule
        Arm & Own optimizer & AdamW & AdamW \(-\) own \\
        \midrule
        \(d=1\) & 70.92 & 72.72 & \(+1.80\pm1.12\) \\
        Per-row \(d=32\) & 71.20 & 72.93 & \(+1.73\pm2.17\) \\
        SPD \(d=32\) & 69.65 & 72.63 & \(+2.98\pm2.39\) \\
        MuonH & 68.56 & 71.66 & \(+3.09\pm2.19\) \\
        Muon & 69.82 & 72.06 & \(+2.24\pm1.33\) \\
        \midrule
        All-\(d=1\), gain LR 0.0048 & 71.45 & 73.03 & \(+1.58\pm1.28\) \\
        Query--key SPD \(d=64\) & 71.90 & 73.43 & \(+1.53\pm1.98\) \\
        Query--key SPD \(d=128\) & 71.15 & 73.21 & \(+2.06\pm1.03\) \\
        \bottomrule
    \end{tabular}
\end{table}

\begin{table}[H]
    \centering
    \small
    \caption{Paired fine-tuned macro-accuracy differences from the matched
    control (percent, 95\% interval, 2 df): no arm beats its control.}
    \label{tab:sft-paired}
    \begin{tabular}{lcc}
        \toprule
        Arm \(-\) control & Own optimizer & AdamW \\
        \midrule
        Per-row \(d=32\) \(-\) \(d=1\) & \(+0.27\) \([-3.36,+3.91]\) & \(+0.21\) \([-2.76,+3.18]\) \\
        SPD \(d=32\) \(-\) \(d=1\) & \(-1.27\) \([-2.67,+0.13]\) & \(-0.08\) \([-2.11,+1.95]\) \\
        MuonH \(-\) \(d=1\) & \(-2.36\) \([-7.15,+2.43]\) & \(-1.06\) \([-3.84,+1.72]\) \\
        Muon \(-\) \(d=1\) & \(-1.10\) \([-1.78,-0.42]\) & \(-0.66\) \([-3.40,+2.09]\) \\
        Query--key SPD \(d=64\) \(-\) all-\(d=1\) & \(+0.45\) \([-2.46,+3.36]\) & \(+0.40\) \([-1.84,+2.64]\) \\
        Query--key SPD \(d=128\) \(-\) all-\(d=1\) & \(-0.30\) \([-2.39,+1.78]\) & \(+0.18\) \([-1.65,+2.01]\) \\
        \bottomrule
    \end{tabular}
\end{table}

No arm beats its control in either column (Table~\ref{tab:sft-paired}). The
only interval that excludes zero is Muon under its own optimizer
(\(-1.10\)); it is one of six comparisons, and Muon has the lowest validation
loss of all arms in that column. Validation loss therefore does not predict
accuracy across arms: the correlation across the eight arm means is 0.02 under
the own optimizer and \(-0.59\) under AdamW. With three seeds, paired
differences vary by about one point, and the study detects effects of about 2
to 6 points, so the intervals bound parameterization effects rather than show
equivalence. AdamW fine-tuning lifts WinoGrande (\(+4.2\) to \(+7.2\)), OBQA
(\(+4.1\) to \(+6.1\)), and SIQA (\(+2.0\) to \(+4.2\)) for every arm.

\section{POET Comparison Details}

OrthAdam \(d=1\) uses a unit-row sphere direction and one learned scalar per
row, with AdamW rather than Muon on the constrained matrix roles.
The POET construction combines two orthogonal factors, a fixed spectrum, a
particular optimization chart, and the numerical policy of the implementation.

\begin{table}[H]
    \centering
    \small
    \caption{One-seed budget sweep for the attention-only controlled POET variant.}
    \begin{tabular}{rrrrr}
        \toprule
        Tokens & POET & OrthAdam \(d=1\) & AdamW & POET--AdamW \\
        \midrule
        0.59B & 3.8418 & 4.2115 & 4.2121 & \(-0.3703\) \\
        2.0B & 3.1019 & 3.1130 & 3.1625 & \(-0.0605\) \\
        6.0B & 2.8430 & 2.8023 & 2.8350 & \(+0.0080\) \\
        30.1B & 2.6660 & 2.6207 & 2.6471 & \(+0.0188\) \\
        \bottomrule
    \end{tabular}
\end{table}

\begin{table}[H]
    \centering
    \small
    \caption{POET's released harness at 130M and 197M tokens, one seed per
    cell. Each entry is the best final loss of a learning-rate sweep run
    separately for each method and precision.}
    \begin{tabular}{lccc}
        \toprule
        Storage & AdamW & POET & POET--AdamW \\
        \midrule
        Full bfloat16 (released) & 3.9427 & 3.9115 & \(-0.0312\) \\
        fp32 & 3.7933 & 3.7846 & \(-0.0087\) \\
        \bottomrule
    \end{tabular}
\end{table}

The released harness casts the full model, including AdamW parameters and
moments, to bfloat16. Standard mixed precision (fp32 parameters and moments,
bfloat16 compute) gives AdamW 3.7854; the released POET kernels do not run
under this mixed-precision path without modification.
The longer-budget table uses a separate attention-only, full-width
implementation. Identity-chart initialization improves its long-budget arm by
0.0111; merge/reinitialize changes it by 0.0005; making the spectrum trainable
improves it by 0.0068. These ablations separate precision, initialization, and
spectrum flexibility.

\section{Pion Comparison Details}
\label{app:pion}

Pion reports final validation loss 2.7350 against 2.7700 for AdamW for a
1.3B model trained on 54B tokens at context length 256, in bfloat16 with the
learning rate 0.0005 selected for AdamW~\citep{shi2026pion}. The released
configuration sets AdamW parameters and ordinary momentum to bfloat16, whereas
Pion's Lie-algebra momentum is created in fp32. We rerun the comparison with
the released code at 1.3B, context length 4096, and \(2\times\) the
Chinchilla budget (48,584 steps, 50.9B C4 tokens, global batch 256, seed 42),
evaluating on C4-val. Every arm uses main learning rate 0.0005 with cosine
decay to 0.000005, \(\beta=(0.9,0.95)\), and weight decay 0.1; AdamW uses 500
warmup steps. The paper configuration of Pion uses \(\beta_2=0.95\),
alternating left and right updates, and a second moment; the repository
default uses \(\beta_2=0.999\), updates both sides at every step, and omits
the second moment. The Pion learning rate applies to the orthogonal updates
only.

\begin{table}[H]
    \centering
    \small
    \caption{Pion at 1.3B, context length 4096, \(2\times\) Chinchilla (one
    seed). Differences are relative to bfloat16 AdamW.}
    \begin{tabular}{lcc}
        \toprule
        Arm & Final loss & \(\Delta\) vs.\ bfloat16 AdamW \\
        \midrule
        AdamW, fp32 & \textbf{2.2655} & \(-0.0804\) \\
        AdamW, bfloat16 (released) & 2.3458 & 0 \\
        Pion, paper configuration, Pion LR 0.00025 & 2.5786 & \(+0.2327\) \\
        Pion, paper configuration, Pion LR 0.0005 & 2.4175 & \(+0.0717\) \\
        Pion, paper configuration, Pion LR 0.001 & 2.3609 & \(+0.0150\) \\
        Pion, paper configuration, Pion LR 0.002 & 2.3741 & \(+0.0282\) \\
        Pion, repository configuration, Pion LR 0.0005 & 2.4312 & \(+0.0854\) \\
        \bottomrule
    \end{tabular}
\end{table}

The paper configuration improves on the repository configuration by 0.0136 at
the same learning rate. The best Pion learning rate, 0.001, is interior: 0.002
finishes at 2.3741, 0.0132 higher. The learning rate 0.001 trails 0.0005 until
step 29,100 and finishes 0.0567 lower, so short learning-rate screens misrank
the two. Even so, Pion remains
0.0150 behind bfloat16 AdamW and 0.0954 behind fp32 AdamW.

\section{StelLA Detailed Results}
\label{app:stella}

\begin{table}[H]
    \centering
    \small
    \caption{StelLA controls at \(S_0=I\) over four paired seeds. Intervals
    are paired 95\% \(t\)-intervals for the difference from Sphere.}
    \begin{tabular}{lcc}
        \toprule
        Method & Accuracy (mean \(\pm\) SD) & \(\Delta\) vs. Sphere \\
        \midrule
        Euclidean & 84.537 \(\pm\) 0.197 & \(-2.116\) \([-2.500,-1.732]\) \\
        Euclidean + factor decay & 86.408 \(\pm\) 0.041 & \(-0.245\) \([-0.589,0.098]\) \\
        Product of spheres & \textbf{86.653} \(\pm\) 0.190 & 0 \\
        Canonical Stiefel & 86.564 \(\pm\) 0.159 & \(-0.089\) \([-0.187,0.009]\) \\
        \bottomrule
    \end{tabular}
\end{table}

\paragraph{Adaptation protocol.}
All runs use StelLA's default \(S_0=I\). The checkpoint selector uses
token-level loss on a fixed 120-example validation split; scores are
macro-averaged over the same eight commonsense tasks as in the main text.

\begin{table}[H]
    \centering
    \small
    \caption{Per-seed StelLA controls at \(S_0=I\). \(^{*}\)The direct-polar
    update is a seed-42 update-rule check.}
    \begin{tabular}{lrrrrr}
        \toprule
        Method & Seed 42 & Seed 43 & Seed 44 & Seed 45 & Mean \(\pm\) SD \\
        \midrule
        Euclidean & 84.405 & 84.814 & 84.387 & 84.543 & 84.537 \(\pm\) 0.197 \\
        Euclidean + U/V decay (1.0) & 86.468 & 86.399 & 86.385 & 86.380 & 86.408 \(\pm\) 0.041 \\
        Product of spheres & 86.514 & 86.616 & 86.553 & 86.930 & 86.653 \(\pm\) 0.190 \\
        Canonical Stiefel & 86.508 & 86.536 & 86.422 & 86.791 & 86.564 \(\pm\) 0.159 \\
        Direct-polar Stiefel\(^{*}\) & 86.641 & -- & -- & -- & -- \\
        \bottomrule
    \end{tabular}
\end{table}

The paired Sphere--Euclidean difference is 2.116 accuracy points
(95\% \(t\)-interval \([1.732,2.500]\)). The Stiefel--Sphere difference is
\(-0.089\) points (\([-0.187,0.009]\)), with Stiefel below Sphere in all four
seeds, and the factor-decay--Sphere difference is \(-0.245\) points
(\([-0.589,0.098]\)).

Direct-polar Stiefel omits only the post-Adam tangent projection and scores
86.641, compared with 86.508 for canonical Stiefel in the same seed-42 check.
The following sensitivity tables come from a separate seed-42 protocol
replication and are not pooled with the paired main-table estimates.

\begin{table}[H]
    \centering
    \small
    \caption{Seed-42 learning-rate sensitivity, scored by eight-task macro
    accuracy. These scores are not used to select the main-table arms.}
    \begin{tabular}{lccc}
        \toprule
        Method & \(10^{-4}\) & \(5\times10^{-4}\) & \(10^{-3}\) \\
        \midrule
        Euclidean, no decay & 84.602 & 84.596 & \textbf{84.777} \\
        Product of spheres & 84.835 & \textbf{86.650} & 86.136 \\
        Canonical Stiefel & 84.835 & \textbf{86.690} & 86.256 \\
        \bottomrule
    \end{tabular}
\end{table}

Sphere and Stiefel attain their best tested value at the default
\(5\times10^{-4}\). Euclidean is best at the upper grid edge \(10^{-3}\) and
remains 1.87 points below Sphere.

\begin{table}[H]
    \centering
    \small
    \caption{Seed-42 factor-wise U/V-decay sensitivity at learning rate
    \(5\times10^{-4}\), scored by eight-task macro accuracy. The middle matrix
    \(S\) has zero decay.}
    \begin{tabular}{rc}
        \toprule
        U/V weight decay & Accuracy \\
        \midrule
        0 & 84.596 \\
        0.01 & 84.542 \\
        0.03 & 85.187 \\
        0.1 & 85.148 \\
        0.3 & 86.250 \\
        1.0 & 86.732 \\
        \bottomrule
    \end{tabular}
\end{table}

The largest tested U/V decay, 1.0, comes within 0.25 points of Sphere in the
four-seed comparison (95\% paired interval \([-0.59,0.10]\)), recovering
about 88\% of the Euclidean-to-Sphere gain. The learning-rate and decay tables
are post-hoc sensitivity checks; main-table checkpoints are selected by loss
on the fixed 120-example validation split.

\section{PSOFT Detailed Results}
\label{app:psoft}

\paragraph{Protocol.}
All arms adapt Llama-3.1-8B on Commonsense-15K with adapters on the query and
value matrices, rank 424, and three epochs, and they share PSOFT's selected
principal subspace, effective initialization, and learned magnitude vectors
\(\alpha,\beta\)~\citep{wu2026psoft}. They differ only in the latent
\(424\times424\) transformation \(R\): PSOFT's five-term Cayley/Neumann
transformation, an unconstrained dense \(R\), or a dense \(R\) whose rows are
initialized at unit norm and kept on unit spheres by tangent projection of the
AdamW update followed by row renormalization. Scores are equal-weight macro
accuracy over the same eight commonsense tasks as in the StelLA study.
Learning rates are selected by the mean best validation loss over seeds 42 and
43: \(8\times10^{-4}\) for PSOFT and \(4\times10^{-4}\) for row-sphere. The
dense arm uses \(4\times10^{-4}\), the validation-best rate of a seed-42 sweep.

\begin{table}[H]
    \centering
    \small
    \caption{PSOFT controls by learning rate and seed, scored by eight-task
    macro accuracy. Bold marks the validation-selected rate of each arm.}
    \begin{tabular}{llrrrr}
        \toprule
        Latent transformation & LR & Seed 42 & Seed 43 & Seed 44 & Mean \\
        \midrule
        PSOFT (Cayley/Neumann) & \(4\times10^{-4}\) & 81.371 & 81.488 & 80.933 & 81.264 \\
         & \(\mathbf{8\times10^{-4}}\) & 80.503 & 80.444 & 80.694 & 80.547 \\
        Dense & \(\mathbf{4\times10^{-4}}\) & 81.047 & 80.984 & 80.828 & 80.953 \\
         & \(8\times10^{-4}\) & 81.439 & 81.237 & 80.936 & 81.204 \\
        Row-sphere & \(\mathbf{4\times10^{-4}}\) & 81.035 & 81.287 & 80.891 & 81.071 \\
         & \(8\times10^{-4}\) & 81.264 & 81.475 & 80.829 & 81.189 \\
        \bottomrule
    \end{tabular}
\end{table}

At the selected rates, row-sphere exceeds PSOFT by 0.52 points, higher in all
three paired seeds, and exceeds the dense arm by 0.12 points, with per-seed
differences \(-0.01\), \(+0.30\), and \(+0.06\). At a shared
\(8\times10^{-4}\), the dense arm matches row-sphere (81.20 versus 81.19),
so the sphere constraint gives no measurable gain there. At a shared learning rate the
ordering of PSOFT and row-sphere reverses: PSOFT leads by 0.19 points at
\(4\times10^{-4}\) and trails by 0.64 at \(8\times10^{-4}\), in each case in all
three seeds. The comparison therefore supports no accuracy cost from replacing
the orthogonal transformation with independently normalized rows, but not a
learning-rate-independent advantage in either direction.

\paragraph{Geometry.}
We inspect the final seed-42 adapters at \(4\times10^{-4}\), 64 matrices of size
\(424\times424\) per arm. All 27{,}136 row-sphere rows have unit norm to within
\(1.2\times10^{-7}\). Dense row norms drift from their unit initialization by
0.006 on average and at most 0.165. Reconstructing PSOFT's five-term Neumann
transformation gives \(\lVert RR^\top-I\rVert_F/\sqrt{424}\) of 0.010 on
average and at most 0.025, with singular values between 0.864 and 1.000. The
implemented transformation is therefore approximately rather than exactly
orthogonal.

\section{Limitations}
\label{app:limitations}

The matched ladder covers 360M models at four budgets and reports C4
validation loss. Every pretraining cell uses three seeds, except the
\(16\times\) run (one) and 1.3B \(d=1024\) SPD (two converged seeds). The
AdamW learning rate for embedding/output-head parameters, normalization
weights, and gains is shared rather than tuned across the Muon-family arms, so
comparisons within the ladder are matched, whereas absolute gaps to Muon and
MuonH are conditional on this setting; tuning only the gain learning rate
improves \(d=1\) by up to 1.8 milli-loss (Appendix~\ref{app:gain-lr}). The
pretraining study uses fixed contiguous row blocks and reports validation loss;
downstream accuracy is measured only after fine-tuning the 3.1B models on one
commonsense-reasoning mixture, where three seeds bound parameterization effects
to a few points; module-wise substitution is evaluated at
360M, and single-projection SPD factors at 360M and 1.3B, with queries and
keys also at 3.1B; learned groupings define complementary
interventions~\citep{yoshihara2026manifold}. The adaptation studies cover
8B-class Llama models on commonsense reasoning.
The image-generation study covers one 130M diffusion transformer on ImageNet at
\(256\times256\).

\end{document}